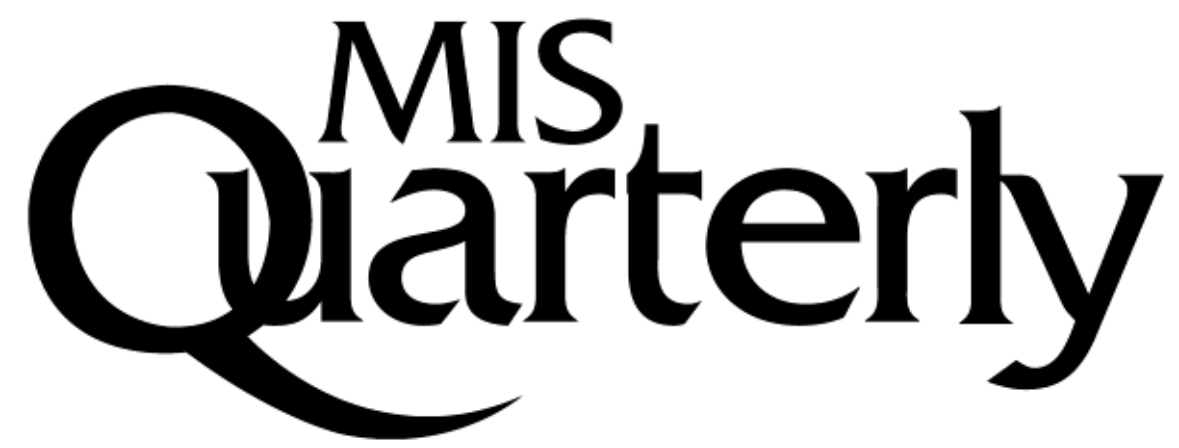

# SHAPLEY VALUE-BASED FEATURE ATTRIBUTION FOR DATA MASKING[1]

**Xinxue (Shawn) Qu and Francis Bilson Darku**
Department of Information Technology, Analytics, and Operations, Mendoza College of Business
University of Notre Dame, South Bend, IN, U.S.A. {xqu2@nd.edu} {fbilsond@nd.edu}

**Hong Guo**
Department of Information Systems, W. P. Carey School of Business
Arizona State University, Tempe, AZ, U.S.A.{hguo@asu.edu}

*Despite its many benefits, widespread access to individuals' personal data also causes severe privacy concerns for consumers, companies, and policymakers. This study proposes a novel framework that adapts the Shapley value-based feature attribution approach to the problem domain of data privacy by capturing the two crucial dimensions of data privacy—disclosure risk and data utility. Our proposed framework takes a holistic view of data masking through a fair feature attribution approach based on Shapley values. Different from the existing literature that mostly focuses on the risk-utility trade-off at the dataset level, the proposed framework addresses the trade-off at the feature level. Furthermore, the proposed framework is agnostic to data masking methods, statistical and machine learning methods, and data utility and disclosure risk evaluation metrics. Experimental results show that our proposed method can effectively reduce disclosure risk while preserving data utility.*

## Introduction

Data has become the most valuable resource in the world (The Economist, 2017). The growth of data resources provides unprecedented opportunities for companies (Aaser & McElhaney, 2021), and data-enabled learning helps create competitive advantages (Hagiu & Wright, 2020). The advent of online tracking and profiling enabled by modern information technology has elevated the importance of data privacy research and practice (Pavlou, 2011). Data privacy commonly refers to protecting individually identifiable personal information and empowering individuals in making their own decisions about who can access/process their data and for what purpose (Smith et al., 2011; Wolford, 2019).

Despite its many benefits, widespread access to individuals' personal data also causes severe privacy concerns for consumers, firms, and regulators. From the consumer perspective, data breaches compromise individuals' sensitive personal information. For example, automakers like General Motors, Ford, and Subaru, in partnership with data brokers such as LexisNexis and Verisk, share consumers' driving data with insurance companies (Hill, 2024). While potentially promoting safer driving, this can raise privacy concerns among policymakers. According to Pew Research (Brooke et al., 2019), the majority of Americans feel a lack of control over data collected about them and are concerned about the use of personal data by firms and the government. In addition to the direct disclosure of private information, inferential disclosure poses a significant risk to consumer privacy protection. On ride-hailing platforms, the collection of riders' and drivers' private information by service providers could lead to inferential disclosure risks. For example, adversaries could potentially monitor drivers' daily routines and infer their home addresses (Zhao et al., 2019). Sensitive information about ride-hailing services, such as the number of rides and utilization of cars, could also be inferred (Pham et al., 2017). From the firm perspective, a study from 2020 calculated potential losses from a data privacy crisis at $79 million, demonstrating the need for companies to update their privacy policies.[2] For example, in a recent iOS software update, Apple introduced a new App Tracking Transparency feature, which requires developers to ask for users' permission when tracking their personal data (Kelly, 2021). In a similar vein, Google stopped tracking individuals' browsing history across multiple websites for the purpose of selling ads (Schechner & Hagey, 2021). Policymakers are also making strides in regulations on data privacy, which specify consumers' privacy rights and firms' privacy obligations. Two prominent examples of such data privacy laws include the General Data Protection Regulation (GDPR) in the European Union and the California Consumer Privacy Act (CCPA) in California.

[1] Balaji Padmanabhan was the accepting senior editor for this paper. Jingjing Li served as the associate editor.

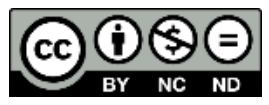

## Problem Setting and Adversarial Model

The data privacy protection issue involves multiple stakeholders, as illustrated in Figure 1, including regulators, consumers/users, firms/organizations, and internal/external data analysts. Catering to increasing demand for organizational collaboration in improving data-driven decision-making processes, firms or organizations often release microdata (i.e., a dataset that contains individual records) to outside data analysts. For instance, firms sometimes collaborate with crowdsourcing platforms (e.g., Kaggle) and provide their proprietary data for machine learning competitions. External data analysts with approved access to the published microdata train statistical and machine learning models to further improve the performance of the model. In a recent competition, American Express turned to the machine learning experts on Kaggle to improve its credit default prediction model by sharing aggregated consumer profile features.[3] Collaboration with external data analysts allows data-owner firms and organizations to obtain improved machine learning models and achieve high data utility.

While collaborative machine learning model development can increase the utility of the shared dataset, sharing access to proprietary data with external analysts can expose consumers and users to disclosure risks, even with nondisclosure agreements (NDAs) and restricted data usage agreements (DUAs) in place (Anand & Lee, 2023). Sharing microdata may expose individuals' confidential information to potential disclosure attacks by malicious intruders, who might directly access or infer confidential information from microdata (Skinner & Elliot, 2002). The possibility of an adversary being able to discover or predict the confidential values of an individual record is known as value disclosure risk (Jiang et al., 2022).

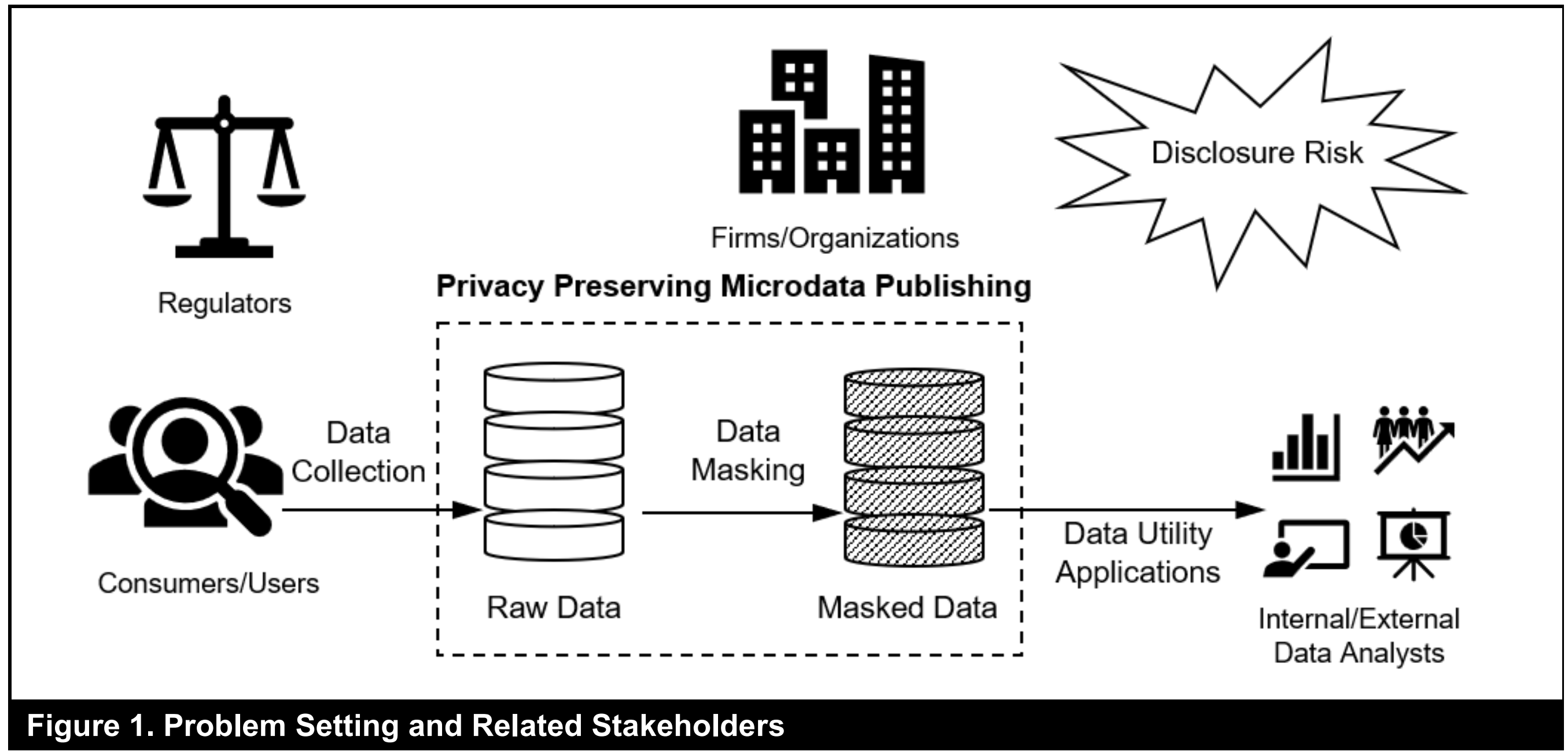

**Figure 1. Problem Setting and Related Stakeholders**

[2] https://www.ftitechnology.com/l/corporate-data-privacy-today

[3] https://www.kaggle.com/competitions/amex-default-prediction/overview

In practice, microdata can be released for the purpose of flexible and effective data analysis, but only after de-identification (i.e., removal of identity-related features such as name and social security number). To further protect individual information from disclosure, confidential attributes are usually preprocessed before being shared.[4] In extreme cases, the data-owner firm may completely remove confidential attributes from the dataset. However, data removal can be detrimental to the utility of the dataset in analytics applications, and research has found that blocking all possible disclosure attacks is untenable (Xu & Zhang, 2022). Therefore, data masking is widely applied to protect confidential attributes. As the firm exerts greater efforts in masking the confidential attribute, the masking mechanism will resemble a scenario where confidential information is removed from the data. Following the literature, this study considers the general scenario under which confidential attributes are masked before being included in the shared dataset.

While masking confidential attributes can mitigate the risk of direct value disclosure, it does not fully address the challenge of inferential disclosure risk. For example, a joint investigation by *The Guardian* and the Danish Broadcasting Corporation revealed that Facebook can infer personal information, such as users' religious and political interests, and share it with advertisers for targeted advertising (Hern, 2018). Similarly, research on ride-hailing services has shown that attackers can monitor drivers' daily routines to infer their home addresses and sensitive business information (Zhao et al., 2019). Once microdata is published, adversaries can utilize the other attributes available in the shared dataset to train a statistical or machine learning model to infer the value of confidential attributes. A well-calibrated prediction model not only enables adversaries to predict confidential values of individual records in shared microdata but may also allow them to infer confidential information for any identified individuals collected from other data sources. In other words, identified individuals whose profiles fit the prediction model well are also subject to high inferential disclosure risks.

Under inferential disclosure risks, Facebook or online advertisers cannot directly observe users' interests, nor can attackers directly collect Uber drivers' home addresses. This is akin to a scenario where the data-owner firm completely removes confidential attributes from the shared dataset. However, the above research demonstrates that these actors were able to collect related information for a subsample and train a model to infer values of the confidential attributes for the rest of the users in the entire sample. Thus, removing or masking confidential attributes in microdata is insufficient to protect an individual's information from inferential disclosure risks. Therefore, it is prudent to also incorporate nonconfidential attributes into the masking procedure. In a general data-sharing scenario, after de-identification, data masking can be employed to reduce the risk of inferential disclosure. This study aims to mitigate inferential disclosure risks through the effective masking of both confidential and nonconfidential features. It contributes to the literature and practical applications by proposing a data masking framework that effectively evaluates feature contributions to disclosure risk and data utility, and recommending efficient feature selection methods for data masking.

## *Research Questions and Contributions*

To reduce inferential disclosure risk, it is important to determine what features should be masked, taking into account the trade-off between risk and utility. In response to this general objective, this paper takes a novel feature attribution approach by posing and addressing the following questions: *How can the contribution of individual features (including both confidential and nonconfidential) be quantified in inferential disclosure risk and data utility? Based on the attribution results, how should features be chosen, and the noise levels for each feature be determined for masking?*

This paper proposes a Shapley value-based feature attribution framework that considers both confidential and nonconfidential features for data masking. Experimental results highlight the importance of masking nonconfidential features, as they might contribute more to inferential disclosure risks than masked confidential features (Ayday & Humbert, 2017; Yang et al., 2019). Our proposed method can effectively reduce disclosure risk while preserving data utility.

The main contributions of this research are threefold. First, we adopt a *holistic view of data masking*. The proposed Shapley value-based feature attribution approach fairly evaluates the contribution of all features in a set of microdata to disclosure risk and data utility, which can then be used to better protect the dataset while maintaining data utility. Second, this work *addresses the risk-utility trade-off at the feature level through selective data masking*. The proposed framework captures two crucial dimensions of data privacy—disclosure risk and data utility, traditionally studied only at the dataset level in the literature. In contrast to existing methods, we quantify each feature's contribution and address the risk-utility trade-off at the feature level. Based on their risk-utility contributions, features are selected for masking to reduce disclosure risk while preserving data utility. Moreover, we introduce the risk-utility frontier to identify candidate features for masking and propose two feature selection criteria, namely risk only and weighted cost. Finally, this research *introduces a model-agnostic data masking framework*. The proposed framework is agnostic to data masking methods, statistical and machine learning models, and metrics for evaluating data utility and disclosure risk. It supports common data masking methods, and the application of Shapley value-based feature attribution is flexible with data utility and inferential models, and works under a variety of evaluation metrics and characteristic functions.

[4] Confidential attributes refer to sensitive private information that an individual typically does not want to be revealed, with examples including income, medical test result, and political orientation.

The rest of the paper is organized as follows. We first review the relevant literature on the topic and then offer a detailed description of the research problem illustrated through an example. Next, we present the Shapley value-based feature attribution approach and introduce our model-agnostic framework for data masking. We then evaluate this framework under different masking approaches and analytics methods. Finally, we conclude with a discussion of our findings and outline potential directions for future research.

## Literature Review

This research is related to the literature on *data privacy* in terms of the problem domain and to the literature on *attribution in machine learning* in terms of research methodology. Figure 2 provides a summary of the related literature and illustrates the position of this study. In this section, we first review the two related literature streams and then discuss the contribution of this paper.

### Data Privacy

The literature organizes potential risks imposed on data privacy into two types: *identity disclosure risk* and *value disclosure risk* (Li & Sarkar, 2014). Identity disclosure risk concerns whether an intruder can link the identity of an individual with a set of feature values, while value disclosure risk concerns whether an intruder can infer the value of a confidential feature in a particular record (Jiang et al., 2022; Li & Sarkar, 2014). This study focuses on the inferential disclosure risk of the confidential feature in a dataset, which can be measured by an intruder's ability to infer the value of a confidential feature for a particular record (Muralidhar & Sarathy, 2006).

To mitigate identity disclosure risks in re-identification attacks, concepts of $k$-anonymity (Gangarde et al., 2021), $l$-diversity (Machanavajjhala et al., 2007), and $t$-closeness (Gangarde et al., 2021) have been introduced to enhance privacy-preserving data applications. De-identification methods, such as generalization (LeFevre et al., 2005), suppression (Kisilevich et al., 2010), and aggregation (Li et al., 2023b), have also proven effective in mitigating identity disclosure risks.

To mitigate inferential disclosure risk, prior studies have proposed different data masking approaches, including *noise insertion*, *micro-aggregation*, *data swapping*, and *data shuffling*. For noise-insertion techniques, values of quantitative features in the original dataset are usually altered by a random noise generated from a predetermined statistical distribution (Brackenbury et al., 2020; Xu & Zhang, 2022). Micro-aggregation approaches mask data by aggregating values of the confidential feature (Wu et al., 2019). Data swapping methods exchange the values of confidential features among individual records that are similar across other feature dimensions (Rodriguez-Garcia et al., 2019). Another related approach is data shuffling, in which the order of original values of the confidential feature is changed, following a conditional distribution approach (Feldman et al., 2022). In addition, synthetic data generation with generative network models has emerged as a promising approach (Xu et al., 2019).

Among all data masking techniques, noise-insertion approaches are the most widely adopted ones, especially in the context of microdata release (Liu et al., 2022). Noise insertion offers a quantified measure of noise and enhanced statistical control over utility losses in data masking. Recently, *differential privacy* (DP) has emerged as a popular approach that provides a privacy loss budget concept and allows data owners to evaluate disclosure risk and determine the amount of noise to add to the response without assuming the intruder's background knowledge (e.g., Li et al., 2023a). Although DP has recently been extended to microdata release (Kenny et al., 2021), setting a privacy budget in DP does not necessarily guarantee the confidentiality of the microdata, and different methods to enforce DP can lead to very different utility and confidentiality levels (Muralidhar et al., 2020). Moreover, the release of microdata contradicts the core idea of DP that the presence or absence of an individual should not lead to changes in the DP output (Domingo-Ferrer et al., 2021). Furthermore, the application of DP to protect large-scale microdata can lead to an unreasonable amount of noise being inserted and a substantial reduction in the utility of the microdata. Our proposed framework is similar to DP in the sense that noise is inserted into the dataset for privacy protection. In contrast to DP, our proposed framework incorporates the risk-utility trade-off into the masking procedure and efficiently allocates noise among features.

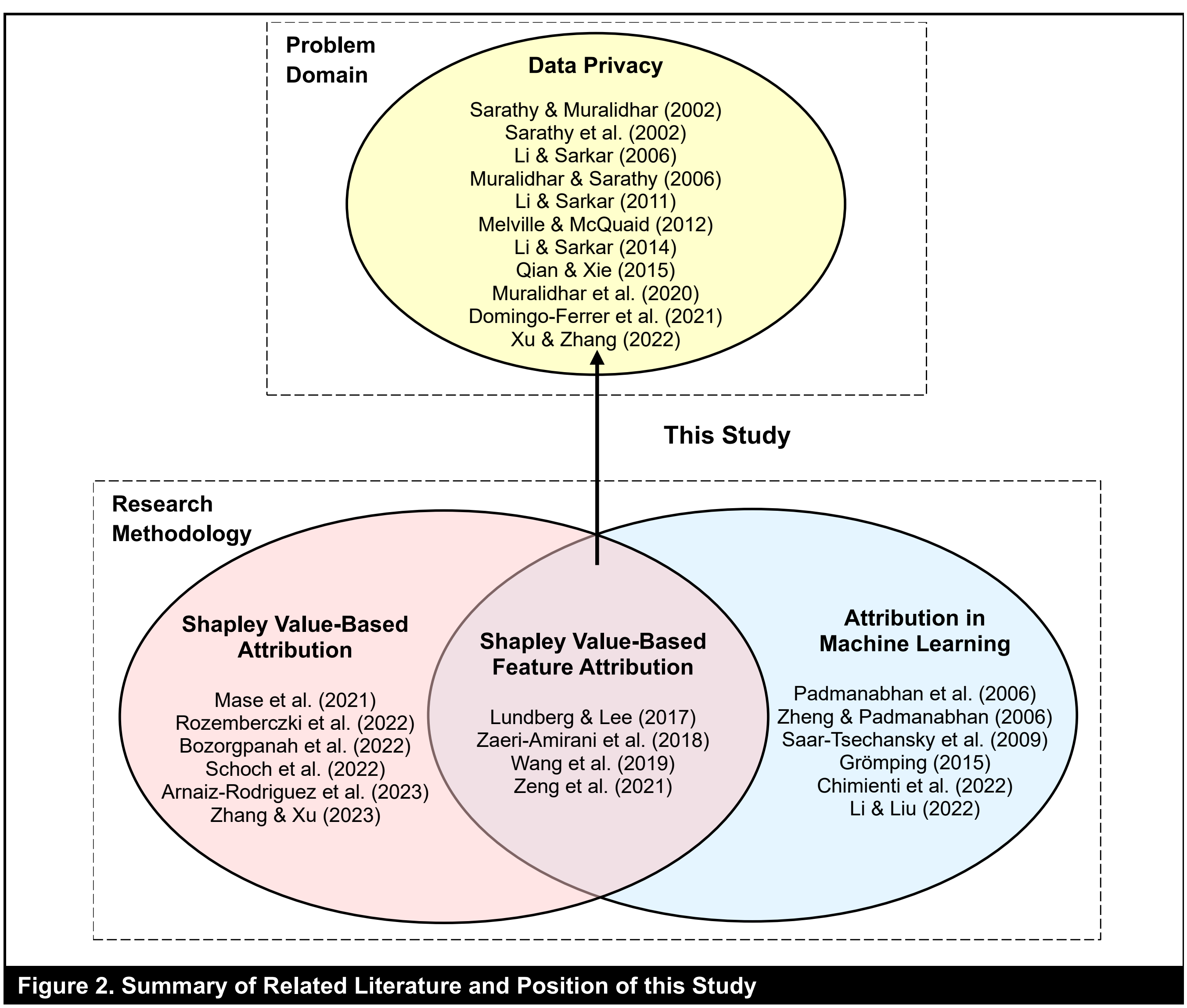

**Figure 2. Summary of Related Literature and Position of this Study**

Masking a confidential feature using the techniques mentioned above can protect its direct disclosure. Nevertheless, the confidential feature is still subject to inferential disclosure. There is a notable similarity between the data privacy and algorithmic fairness fields regarding the challenges posed by the correlation between confidential (protected) and nonconfidential (nonprotected) features. As nonconfidential features can contribute to disclosure risk, nonprotected features can also lead to unfair model outcomes because unobserved protected attributes may have an indirect influence on the model outcome (Pan et al., 2021). Fairness research has utilized the relationship between protected and nonprotected attributes to impute unobserved protected attributes (Kallus et al., 2022). In contrast, data privacy studies examine how intruders can be prevented from exploiting relationships among features to infer confidential information.

While some studies have recognized the potential risks of nonconfidential features in inferential disclosure (Ayday & Humbert, 2017; Yang et al., 2019), a systematic approach to incorporating nonconfidential features in data privacy protection is lacking. Although Li and Sarkar (2014) proposed a method based on nonconfidential features to protect against regression attacks on confidential features, in their experiments, either all or half (randomly selected) of the nonconfidential features in the dataset were masked. Masking all nonconfidential features would lead to a potential loss in data utility without a corresponding decrease in the disclosure risk. This introduces the challenge of how to select a subset of nonconfidential features to protect data privacy while preserving data utility. This study addresses this challenge by systematically exploring the role of nonconfidential features in data privacy protection. Specifically, we apply a Shapley value-based feature attribution approach to quantify the contributions of individual features to inferential disclosure risk and data utility and suggest feature selection criteria for masking.

Data masking techniques can reduce the risk of disclosure; however, masking applied to datasets can tweak the statistical characteristics of the dataset and even distort the relationship among features. Data masking can lead to changes in the statistical properties of the dataset, potentially introducing biases (Muralidhar et al., 1999) or affecting the relationships among features (Sarathy & Muralidhar, 2002). Although the trade-off between disclosure risk and data utility has been recognized and applied to evaluate the performance of different data masking methods, existing studies mainly address this trade-off at the dataset level (Li & Sarkar, 2011; Melville & McQuaid, 2012). Because risk/utility is measured at the dataset level, a systematic evaluation of how each feature contributes to the disclosure risk and data utility is lacking. Therefore, the existing data masking techniques cannot effectively determine which features to mask and how much noise to insert for the selected features. Moreover, the inferential disclosure risk is essentially caused by feature-level correlation between confidential features and nonconfidential features. As a result, it is important to understand the contribution of individual features to the inferential disclosure risk. This paper complements this stream of research by addressing the risk-utility trade-off at the feature level using a Shapley value-based feature attribution approach.

Although some data masking algorithms can preserve the distribution of the confidential feature (Qian & Xie, 2015) or correlation between features (Sarathy et al., 2002), these methods cannot be easily adapted to general applications with different data utility metrics. In contrast, the framework proposed in this paper is agnostic to the various data utility metrics, data masking methods, statistical and machine learning models, and disclosure risk metrics.

## Attribution in Machine Learning

This study is related to the broader research stream of attribution in machine learning that quantifies and attributes overall utilities to individual (or groups of) features, data records, or ensembling models. The attribution problem has been studied in a wide range of research contexts, including feature selection (Solorio-Fernández et al., 2020), data valuation (Ghorbani & Zou, 2019), and machine learning explainability (Burkart & Huber, 2021). In the context of data valuation, firms need to evaluate the related data acquisition cost and benefits of modeling improvement (e.g., Padmanabhan et al., 2006). Zheng and Padmanabhan (2006) optimized a dual objective that takes into account both global model performance and subsequent imputation accuracy, and Saar-Tsechansky et al. (2009) proposed an acquisition policy based on estimates of the expected utility from potential acquisitions. In a similar vein, some recent studies have investigated the contribution of individual sample (groups) to machine learning model performance (Chimienti et al., 2022) and treatment fairness (Li & Liu, 2022).

From the perspective of research methodology, this paper closely builds on the literature on feature attribution, particularly the application of Shapley values in evaluating how much each feature contributes to the performance of a model for a given metric. In statistics, the discussion of feature attribution mainly takes place in the context of finding the proportionate contribution of each feature in a regression model setting (Grömping, 2015). In computer science, feature attribution has also been used to assign the contribution of features in machine learning algorithms (Lundberg & Lee, 2017). However, the literature on feature attribution has mainly focused on the decomposition and distribution of $R^2$ among the independent features in a (regression) model (see Grömping, 2015 for a review). Thus, existing methods have not been generalized to different model evaluation metrics (e.g., prediction accuracy, root mean square error).

The Shapley value, proposed by and named after Lloyd Stowell Shapley (1953), is a well-known concept in cooperative game theory that distributes the total surplus generated by a coalition to each player within the coalition. The Shapley value approach is an efficient solution concept that offers complete decomposition to the attribution problem and provides fair evaluations of player contributions. Moreover, the Shapley value approach is model agnostic and can be used with any characteristic function, making it a versatile tool for feature attribution in many applications.

Given the merits of the Shapley value concept, it has been widely adopted as the state-of-the-art method in machine learning and other fields to, for example, explain how features contribute to the preditions of classification and regression models (Lundberg & Lee, 2017; Zaeri-Amirani et al., 2018), provide valuations of the contributions of training instances (Schoch et al., 2022), quantify the importance of individual models in ensemble games (Rozemberczki & Sarkar, 2021), and evaluate the importance of grouped features in federated learning (Wang et al., 2019; Zeng et al., 2021). In regression analysis, the use of Shapley values to measure the contribution of each independent feature to the proportion of variance in the dependent feature explained by the model is commonly known as Shapley regression (Lipovetsky, 2006). In the related data privacy literature, Bozorgpanah et al. (2022) utilized Shapley values as a metric for data utility to assess the effects of data masking on data utility. In the emerging field of algorithmic fairness, extant studies have applied the Shapley value approach to evaluate the contribution of features (e.g., Pan et al., 2021) and data records (e.g., Arnaiz-Rodriguez & Oliver, 2023) to a predefined fairness metric. For example, the Shapley value approach was applied to evaluate the extent of bias to which each individual in a dataset was adversely or beneficially subjected due to protected attributes (Mase et al., 2021). For interested readers, Rozemberczki et al.'s (2022) survey paper provides a comprehensive review of the application of the Shapley value in machine learning.

Overall, as depicted in Figure 2, this study proposes a novel framework that adapts the Shapley value-based feature attribution approach to the problem domain of data privacy. Specifically, we modified the Shapley value approach to capture the two crucial dimensions of data privacy—disclosure risk and data utility. Based on their two-dimensional Shapley values, features may be selected for masking to balance the risk-utility trade-off. The contribution of this paper relative to the existing literature is threefold. First, the proposed framework takes a holistic view of data masking and is the first to utilize the Shapley value-based feature attribution approach to provide a fair evaluation of how individual features, both confidential and nonconfidential, contribute to disclosure risk and data utility. Second, the existing literature on data privacy mostly focuses on the risk-utility trade-off at the dataset level. This paper addresses the risk-utility trade-off at the feature level by identifying candidate features for masking using the risk-utility frontier and selecting specific features for masking based on two criteria: risk only and weighted cost. Lastly, the proposed framework is model agnostic and can thus be applied with different masking approaches, predictive/inferential models, and various metrics commonly considered in the literature.

## Problem Description and an Illustrative Example

Given the pressing concerns about data privacy, policymakers have issued data privacy regulations, including general data privacy laws such as the GDPR in the European Union and the CCPA in California, as well as industry-specific data regulations such as the Personal Financial Data Rights Rule and the Health Insurance Portability and Accountability Act (HIPAA). Major data privacy regulations strictly limit the information firms and organizations can collect from consumers when constructing microdata. For example, following the enactment of the GDPR and CCPA, organizations are required to disclose their data collection mechanisms to consumers, limit their use of consumers' information to the disclosed purposes, and obtain consumers' consent on the collection and sharing of related features.

For microdata collected in compliance with current regulations, the data owner must implement de-identification to protect individuals' identities. For instance, to satisfy the de-identification standard of HIPAA, the Safe Harbor approach requires the elimination of 18 personal identifiers, including SSN, email address, and IP address, before secondary use of data for policy assessment, life sciences research, and other endeavors (Office for Civil Rights, 2012). After de-identification, the dataset has to be properly masked before sharing. The data owner can determine the sample size and the amount of noise to insert into the masking process (Anand & Lee, 2023). This paper aims to improve the data masking process before data sharing by providing an efficient and effective feature evaluation and selection framework that can balance the trade-off of risk and utility at the feature level.

In this section, we first introduce the notations and provide a formal problem description. We then introduce an illustrative example, which will be used as a running example throughout the paper to demonstrate our research problem and the proposed solution.

### Notations and Problem Description

Following convention, we consider a dataset that consists of $p$ features and $n$ individual records. Among the $p$ features, some features are confidential while others are nonconfidential. For ease of illustration, in the main analysis of this study, we focus on the scenario when there is only one confidential feature, $y$, in the dataset. But the proposed framework also applies to scenarios with multiple confidential features in the dataset, which is analyzed in detail in the Application to Scenarios With Multiple Confidential Features subsection. Without loss of generality, suppose the first $p-1$ features (i.e., $X = \{x_1, \dots, x_{p-1}\}$) are nonconfidential features and the last feature is the confidential feature. Thus, we would have a dataset $\mathcal{D}$ with $p$ features, $\mathcal{D} = X \cup \{y\} = \{x_1, x_2, \dots, x_{p-1}, y\}$. To prevent the direct disclosure risk when sharing the dataset $\mathcal{D}$, the confidential feature $y$ needs to be masked. Let $f: y \rightarrow y'$ be any masking technique that protects the confidentiality of $y$ and produces a value $y'$ for sharing purposes. The dataset after masking the confidential feature will be $\mathcal{D}' = X \cup \{y'\} = \{x_1, x_2, \dots, x_{p-1}, y'\}$.

After masking, the dataset can be shared to users who are granted access for data utility applications. The target feature is represented as feature $z$, which can be inferred or predicted with statistical or machine learning models (denoted as function $h: \mathcal{D}' \rightarrow \hat{z}$ ). The inferred or predicted value of the data utility feature is denoted using $\hat{z}$, for which $\hat{z} = h(x_1, \dots, x_{p-1}, y')$. The utility of the dataset can be evaluated based on the inferential performance metrics (e.g., $R^2$, MSE, RMSE) and predictive performance metrics (e.g., precision, recall, ROC) of the application tasks.

As the dataset is shared for analytics applications, the confidential feature $y$ is subject to potential inferential disclosure risks. Potential intruders could take advantage of various statistical inference or machine learning models, denoted as function $g: \mathcal{D}' \rightarrow \hat{y}$, to infer the values of the confidential feature. The inferred confidential value is $\hat{y}$, where $\hat{y} = g(x_1, \dots, x_{p-1}, y')$. Then, the inferential disclosure risk can be measured based on the relationship between the actual value, $y$, and the inferred value, $\hat{y}$, of the confidential feature using distance or variance metrics. Table 1 summarizes the key notations used in this paper.

Masking the confidential feature alone cannot fully prevent disclosure attacks, as adversaries can use nonconfidential features in the same dataset to infer confidential features. A naive approach would be to mask all the features in the dataset, but this could drastically reduce data utility without effectively reducing the inferential disclosure risk. Moreover, different features would pose different inferential disclosure risk levels. Thus, it is important to identify and prioritize high-risk features for masking (Li & Sarkar, 2014). This study focuses on two main questions: (1) *How can the contribution of individual features (including both confidential and nonconfidential) to both inferential disclosure risk and data utility be quantified?* (2) *Which features should be masked, and how much noise should be inserted in the masking of the selected features?*

## An Illustrative Example

Consider a dataset containing individual consumers' information collected by a bank. This dataset classifies consumers described by a particular set of features as either good or bad credit risks (Hofmann, 1994). A description of the full dataset can be found in the Experimental Studies section. For our illustration, we use a subset of 200 randomly selected consumer records with seven important quantitative features as a running example (shown in Table 2).

In this example dataset, some features are demographic or economic in nature, such as age (*Age*) and length of current residence (*Residence*). Other features are banking and credit related and include the number of existing credit lines at the bank (*NumCredit*), the credit duration in months (*Duration*), and the number of people the bank is liable to provide maintenance for (*NumLiable*). The feature *CreditAmount* is considered to be the confidential feature that needs to be protected.

**Table 1. Summary of Notation**

| Notation | Description |
|---|---|
| $i$ | Subscript $i$ is the index for the individual feature |
| $j$ | Subscript $j$ is the index for the individual data record |
| $\mathcal{D}, \mathcal{D}'$ | Original and masked datasets respectively |
| $p$ | Number of features in one dataset |
| $x_i, i \in \{1, 2, \dots, p-1\}$ | Nonconfidential feature |
| $y, y'$ | Confidential and masked confidential features, respectively |
| $\hat{y}$ | Predicted confidential feature |
| $z, \hat{z}$ | Actual and predicted utility features respectively |
| $f(\cdot)$ | Data masking function |
| $g(\cdot)$ | Privacy inferential model |
| $h(\cdot)$ | Data utility model |
| $r(\cdot)$ | Risk metric |
| $u(\cdot)$ | Utility metric |
| $\phi_i(r)$ | Shapley value of risk for feature $i$ |
| $\phi_i(u)$ | Shapley value of utility for feature $i$ |
| $\mathcal{F}$ | Risk-utility frontier |
| $t(\cdot,\cdot)$ | Feature selection criterion |

**Table 2. Illustrative Example: Credit Data**

| No. | Age $x_1$ | NumLiable $x_2$ | NumCredit $x_3$ | Residence $x_4$ | Duration $x_5$ | CreditAmount $y$ | Credit_Mask $y'$ | Credit_Infer $\hat{y}$ |
|---|---|---|---|---|---|---|---|---|
| 1 | 63 | 1 | 1 | 4 | 15 | 1520 | 2064.55 | 1920.16 |
| 2 | 28 | 1 | 1 | 2 | 24 | 1249 | 3417.74 | 2883.15 |
| 3 | 68 | 1 | 1 | 4 | 6 | 14896 | 14811.35 | 7577.61 |
| ⋮ | ⋮ | ⋮ | ⋮ | ⋮ | ⋮ | ⋮ | ⋮ | ⋮ |
| 199 | 43 | 1 | 2 | 1 | 24 | 1516 | 1773.93 | 2260.98 |
| 200 | 42 | 1 | 1 | 3 | 18 | 4153 | 2118.79 | 3361.71 |

When the bank releases a dataset for consumer credit evaluation, the confidential feature, *CreditAmount* ($y$), is usually masked to protect confidential values from direct disclosure. The released dataset includes a masked confidential feature, *Credit_Mask* ($y'$). The potential direct disclosure risk can be measured by the distance between the value of the masked confidential feature and its actual value, $r(y, y')$. In the illustrative example, the distance between *CreditAmount* ($y$) and *Credit_Mask* ($y'$), measured as average absolute distance (AAD), is 2641.11. The larger the distance metric, the higher (lower) the level of data privacy (disclosure risk).

As the dataset is released, potential intruders may also gain access to the masked dataset. In addition to inferring the confidential value from the masked confidential feature (*Credit_Mask*), other nonconfidential features may also be included in a model to infer the true value of the confidential feature. For instance, potential intruders can calibrate a simple linear regression model to infer the value of *CreditAmount* using all features in the released dataset, $\hat{y} = g(x_1, \ldots, x_6, y')$.[5] The predicted value is presented as *Credit_Infer* ($\hat{y}$) in Table 2. Such inferential disclosure risk can be measured by the distance between the predicted value and the actual value of the confidential feature. In this example, the distance measured in AAD is 1562.53, which is much smaller than that measured by direct disclosure. The small AAD distance indicates a high accuracy of the intruder's prediction of the confidential information. As a result, in addition to masking the confidential feature, it is also important to understand how nonconfidential features contribute to such inferential disclosure risk. More generally, an effective data masking mechanism at the feature level is needed to provide the dataset with comprehensive protection.

[5] Although potential intruders may not know the actual value of the confidential feature, some unsupervised methods (e.g., principal component analysis) and/or extra surveys can still be conducted.

## Shapley Value-Based Feature Attribution for Data Masking

In this section, we introduce the Shapley value-based feature attribution framework. We seek to address the research questions mentioned in the previous section by proposing a framework for data disclosure risk evaluation, as well as some suggestions on data privacy protection, by incorporating the trade-off between data utility and inferential disclosure risk during data masking. In what follows, we describe in detail the scope and individual components of the proposed framework, as illustrated in Figure 3.

### *The Scope of the Framework*

This study focuses on protecting *structured de-identified* microdata from *inferential disclosure risk*. We assume that the microdata is structured with rows containing observations on individual consumers/users and columns containing features about the individuals in the dataset. The microdata contains no direct identifiers—that is, none of the features can be used to directly identify any of the entities whose records are in the dataset. If such features exist (e.g., names, social security numbers), the microdata will need to be de-identified first by removing those features. In the de-identified microdata, one or more features are determined a priori as confidential, either by regulators or the data owner, and, consequently, the other features will be considered nonconfidential. Although we focus on the scenario with one confidential feature in the main analysis for ease of illustration, the proposed framework can be applied to scenarios with multiple confidential features, which we analyze in the Application to Scenarios With Multiple Confidential Features subsection.

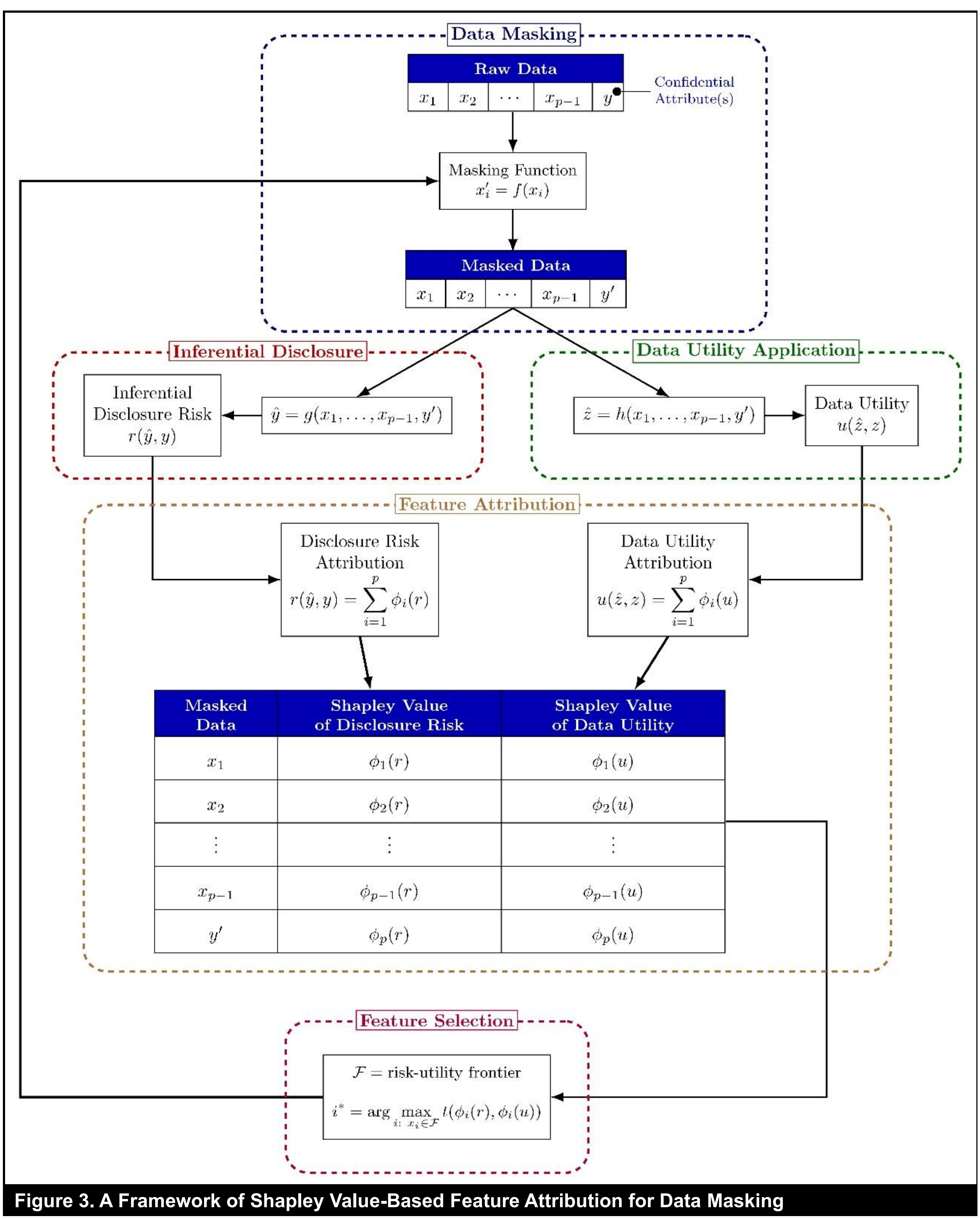

**Figure 3. A Framework of Shapley Value-Based Feature Attribution for Data Masking**

The framework is designed to work with *continuous features* using noise insertion-based data masking techniques. Existing work in data privacy typically proposes different data masking techniques to protect categorical and continuous features from disclosure risk. In the research stream on privacy protection in microdata release, most prior studies designed data masking techniques for continuous features, where better control over the level of perturbation can be achieved through a tuning parameter for inserting noise into the features (e.g., Brackenbury et al., 2020; Kim & Winkler, 2003; Lee et al., 2010; etc.). That being said, we recognize the value of handling categorical features, but leave this to future investigation.

To mitigate inferential disclosure risk, we propose a *feature-level* data masking framework that evaluates each feature's contribution to disclosure risk and data utility and implements data masking at the feature level. Masking the confidential feature, or eliminating it from the microdata, can protect the microdata against direct disclosure risks. However, nonconfidential features can still be exploited to infer confidential information due to their (statistical) relationship with the confidential feature. To understand the contribution of individual features to the inferential disclosure risk and the data utility, the proposed framework applies the Shapley value approach to attribute overall risk and utility to the features. Based on the calculated Shapley values of each feature, the proposed framework identifies which features (masked confidential and unmasked nonconfidential features) contribute more to inferential disclosure risk and how much noise should be inserted while keeping the selected features' statistical properties for data utility.

In summary, this paper focuses on protecting structured de-identified microdata from inferential disclosure risk. The proposed feature-level data masking framework works with continuous features by selecting features based on their contributions to inferential disclosure risk and data utility and providing guidance regarding how much noise to insert into the feature masking process.

## Data Masking

The typical procedure to process de-identified microdata before sharing or publicizing is to mask it, which we denote as the *data masking* process in Figure 3. Traditional methods, such as additive and multiple data perturbation (Muralidhar et al., 1995), general additive data perturbation (Muralidhar et al., 1999), copular-based perturbation (Sarathy & Muralidhar, 2002), multiple imputation with multimodal perturbation (Melville & McQuaid, 2012), and skew-*t* data perturbation (Lee et al., 2010) can be implemented in the masking procedure. In this study, both confidential and nonconfidential features are candidates for the data masking process.

Having the original dataset $\mathcal{D}$ at the initial stage of this framework, the confidential feature $y$ will first be masked into $y'$ using any of the appropriate privacy-preserving methods $f(\cdot)$, i.e., $f: y \rightarrow y'$, with a specified noise level $\delta$. The confidential feature $y$ is then replaced by the masked confidential feature $y'$ in the dataset, yielding the modified dataset $\mathcal{D}'$.

## Inferential Disclosure Risk

As the dataset $\mathcal{D}'$ is published or shared for utility applications, it also becomes exposed to potential intruders. Shared features (including the masked confidential feature $y'$) can be used to infer the true value of confidential feature $y$. The possibility that the confidential features can be inferred by unauthorized users is referred to as *inferential disclosure*. Even though the masked confidential feature $y'$ alone poses some level of disclosure risk, the presence of other features in the released dataset $\mathcal{D}'$ increases that level of risk.

To measure the total disclosure risk of the shared dataset $\mathcal{D}'$, this step of the framework requires the use of a chosen statistical or machine learning model $g(\cdot)$ that treats the confidential feature $y$ as a dependent feature and all the features in $\mathcal{D}'$, including $y'$, as independent features. The model $g(\cdot)$ can be a linear regression model, generalized additive model, regression tree model, or any other appropriate model that can use the dataset to predict the values of $y$. How close the predicted values $\hat{y}$ are to the original feature $y$ is measured as the inferential disclosure risk $r(\mathcal{D}') \equiv r(\hat{y}, y)$ of the dataset $\mathcal{D}'$. The function $r$ can be any desirable inferential or model performance metric, which may include correlation-based metrics (e.g., $R^2$, canonical correlation), distance-based metrics (e.g., mean square error, mean absolute error), or other metrics.

## Data Utility Application

The secondary goal of a good data masking technique $f$ is to preserve the utility of the original dataset $\mathcal{D}$ in the masked dataset $\mathcal{D}'$. This is to ensure that the statistical properties of the dataset are maintained to a certain degree for reasonable inferential statistics and predictive analytics. Thus, once a feature is masked, it is imperative to measure data utility of the shared dataset $\mathcal{D}'$.

In this step of the framework, an appropriate data utility metric $u$ can be chosen to quantify the utility of the entire dataset $\mathcal{D}'$, as opposed to only the masked feature $y'$. The idea of quantifying the utility of the entire dataset stems from the fact that other features aside from y' may be masked in subsequent steps; thus, a metric that captures the total utility of all the features in the dataset will be more suitable.

In data utility applications such as inferential and predictive analyses, the performance of the shared dataset $\mathcal{D}'$ in inferring and predicting the response feature $z$ can be used as a utility metric $u(\mathcal{D}')$. Therefore, in this case, the utility of the dataset $\mathcal{D}'$ will be how close predicted values $\hat{z}$ are to the original feature $z$ $\left(u(\mathcal{D}') \equiv u(\hat{z}, z)\right)$. The function $u$ can be any performance metric for data utility, such as correlation-based or distance-based metrics.

## *Shapley Value-Based Feature Attribution*

The previous two steps measure the disclosure risk and utility at the dataset level. However, features in the shared dataset contribute differently to potential data utility application and disclosure risk. The next step in this framework is the Shapley value-based feature attribution, which fairly evaluates the contribution of features to disclosure risk and data utility.

The Shapley value approach is a widely used attribution method that originated from cooperative game theory. We first assume there is a set of players, $P = \{1,2,3, \dots, p\}$. For any coalition (subset) of players, $S \subseteq P$, the *characteristic function* $v: S \to \mathbb{R}$ describes the value of the coalition. Shapley value provides a fair allocation of the value of the grand coalition $v(P)$ to each player $i \in P$. Shapley (1953) defines the contribution of each player $i$ as

$$\phi_i(v) = \sum_{S \subseteq P \setminus \{i\}} \frac{s!\,(p-s-1)!}{p!} [v(S \cup \{i\}) - v(S)], \qquad (1)$$

where $p$ and $s$ are the cardinalities of $P$ and $S$, respectively. The value $\phi_i(v)$ of each player is the weighted average of the marginal contributions of that player for all possible coalitions it can join.

In data masking, players in a game correspond to features in a dataset, while the characteristic function corresponds to disclosure risk or data utility. The $p-1$ nonconfidential features, $X = \{x_1, x_2, \dots, x_{p-1}\}$, together with the masked confidential feature, $y'$, are *players* in the calculation of Shapley values of disclosure risks and data utility. Let $S$ be a subset of features from the shared dataset $\mathcal{D}'$ ($S \subseteq \mathcal{D}'$), which represents a *coalition* in the game. For any subset of features $S \subseteq \mathcal{D}'$, $r(S) \in \mathbb{R}$ is a performance metric for measuring the inferential disclosure risk of $y$ from a model using a subset of features $S$, and $u(S) \in \mathbb{R}$ is a performance metric for measuring the strength of a predictive model for $z$ from a subset of features $S$. Using Equation (1), $\phi_i(r)$ denotes the contribution of feature $x_i$ (for $1 \le i \le p$, where $x_p \equiv y'$) to the total disclosure risk of the dataset $\mathcal{D}'$ (i.e., $r(\mathcal{D}')$) in exposing the confidential feature $y$. Likewise, the value $\phi_i(u)$ allocates to feature $x_i$ its contribution to the total data utility of the dataset $\mathcal{D}'$ (i.e., $u(\mathcal{D}')$) in predicting $z$.

### Properties and Computation of Shapley Value

Shapley values calculated based on Equation (1) satisfy the following properties:

**Efficiency:** The sum of the Shapley values of the features equals the value of the shared dataset, so that all the risk or utility of the shared dataset is distributed among the features, i.e., $\sum_{i \in P} \phi_i(r) = r(\mathcal{D}')$ and $\sum_{i \in P} \phi_i(u) = u(\mathcal{D}')$.

**Symmetry:** If two features have equal contribution measured by the characteristic function, $r(\cdot)$ (or $u(\cdot)$), for every subset $S$ they join, they should each have equal values assigned to them by the Shapley value $\phi$. That is, if $\forall S \subseteq \mathcal{D}' \setminus \{x_i, x_j\}$, $r(S \cup \{x_i\}) = r(S \cup \{x_j\})$, then $\phi_i(r) = \phi_j(r)$.

**Dummy or null feature:** Any feature that does not contribute to the risk (or utility) of any subset of the shared dataset it is added to should have a Shapley value of 0. In other words, if $\forall S \subseteq \mathcal{D}' \setminus \{x_i\}$, $r(S \cup \{x_i\}) = r(S)$, then $\phi_i(v) = 0$.

For simplicity and ease of interpretation of the Shapley values for risk and utility, we consider characteristic functions $r$ and $u$ that satisfy the following properties. First, the risk and utility of any model that does not use any feature from the dataset $\mathcal{D}'$ is zero, i.e., $r(\emptyset) = 0$ and $u(\emptyset) = 0$. Second, the addition of more features to a model does not decrease the risk or the utility, that is $r(S_1 \cup S_2) \ge r(S_1)$ and $u(S_1 \cup S_2) \ge u(S_1)$, where $S_1, S_2 \subseteq \mathcal{D}$. These mild assumptions about the characteristic functions are commonly made or implied in the literature on Shapley values (Shapley, 1953). Performance metrics of inferential disclosure risk and data utility (e.g., $R^2$, average absolute distance [AAD], average squared distance [ASD], root average squared distance [RASD]) or their transformations that satisfy the two assumptions, can be used as $r(\cdot)$ and $u(\cdot)$ functions. For example, if $R^2$ is chosen as a metric for a linear regression model to measure disclosure risk, then $r(\emptyset) = R^2(\emptyset) = 0$ and its value increases as more predictors are included in the model. If ASD is chosen, then a simple transformation using $r(S) = ASD(\emptyset) - ASD(S)$ will satisfy both conditions. These transformations make the interpretation of the characteristic functions straightforward and consistent.

### Computation of Shapley Value

The calculation of the Shapley value is well-known to be computationally expensive. In fact, it increases by the order of $2^p$, where $p$ is the number of features. To ease the computational burden of obtaining Shapley values, the marginal contribution of each feature to coalitions can be sampled based on a probabilistic characterization of Equation (1). By doing so, the Shapley value $\phi_i$ is redefined as the expected value of the marginal contribution of a feature $i$ to a coalition $S$ (Castro et al., 2017; Han et al., 2021). Thus, Equation (1) can be written as $\phi_i(v) = \mathbb{E}[\Delta v_i(S)] = \sum_{S \subseteq \mathcal{D}' \setminus \{i\}} P(S) \Delta r_i(S)$, where $\Delta v_i(S) = v(S \cup \{i\}) - v(S)$, and $P(S) = \frac{s!(p-s-1)!}{p!}$.

This paper does not focus on which approximation method is efficient, accurate, or faster. Therefore, for a dataset with a relatively small number of features, we apply the exact Shapley value calculations. For a relatively larger dataset with more features, we use the approximation methods of Castro et al. (2017) and Okhrati and Lipani (2021). For a comprehensive review of the Shapley value approximation methods, see Rozemberczki et al. (2022).

### Shapley Value on Risk and Utility Dimensions

Once characteristic functions $r(\cdot)$ and $u(\cdot)$ are selected, Shapley values for risk and utility for each feature can be computed based on Equation (1). Shapley values for risk $\phi_i(r)$ and utility $\phi_i(u)$ of feature $i$, respectively, represent contributions that feature $i$ makes to the inferential disclosure risk and data utility of the shared dataset.

Figure 4 shows both the Shapley value of risk and the Shapley value of utility for all features in the running credit example. As illustrated in Figure 4, the masked confidential feature *Credit_Mask* has the highest Shapley value of risk and the highest Shapley value of utility, indicating that it is contributing the most to both risk and utility. In other words, although masking the confidential feature significantly reduces disclosure risk, it also hurts data utility. A nonconfidential feature *Duration* has the second-highest Shapley value of risk, but a much lower Shapley value of utility, indicating a high contribution to risk but a low contribution to utility. In other words, masking *Duration* will significantly reduce disclosure risk without sacrificing too much data utility. Therefore, *Duration,* in addition to *Credit_Mask,* is also a good candidate for data masking.

Overall, the Shapley value-based feature attribution enables a better understanding of the trade-off between data utility and data disclosure risks at the feature level. Features (confidential or nonconfidential) with a high Shapley value of risk and a low Shapley value of utility should be considered for masking. The next subsection will discuss how to select features for masking.

## *Feature Selection*

Following the proposed Shapley value-based feature attribution approach, the understanding of the risk-utility trade-off can be extended to the feature level. When selecting the next feature to mask in the data masking framework, the calculated Shapley values of risk and utility for features need to be properly incorporated to resolve the conflict between risk and utility.

To effectively reduce the risk level of a dataset while maintaining its utility, features with a high Shapley value of risk but a low Shapley value of utility should be selected to mask. Such features contribute the most to the risk level of the dataset, while masking them may impact the utility of the dataset to a limited extent. When dealing with a high-dimensional dataset, it is computationally expensive to consider all available features in the masking procedure. Based on the Shapley value of each feature in both disclosure risk and data utility dimensions, we define the *risk-utility frontier* consisting of all candidate features as follows:

> ***Risk-utility frontier ($\mathcal{F}$)*** *is a polyline defined by a set of features that contribute the highest disclosure risk for a defined level of data utility or the lowest data utility for a given level of disclosure risk.*

Following the definition, features on the risk-utility frontier[6] are never dominated by other features in the disclosure risk and data utility plane. In other words, masking features that are not on the risk-utility frontier is suboptimal because masking them does not significantly reduce the disclosure risk for the same level of data utility loss. Thus, focusing on features located on this risk-utility frontier improves efficiency in feature selection. As illustrated in Figure 5, features located on the frontier (such as *Credit_Mask*, *Duration*, *Residence*, and *Age*) contribute the highest risk given the same amount of utility. It is important to note that as the masking procedure proceeds, the masked confidential feature shifts away from the risk-utility frontier toward the internal region, suggesting that masking only the confidential feature might not be an efficient solution.

[6] The concept of the risk-utility frontier is an application of *Pareto front* (Varian, 1974) in the data privacy domain, which is similar to *skyline* (Borzsony et al., 2001) in the computer science literature.

Features located on the risk-utility frontier provide a set of candidate features to mask. However, it is not possible to select high-risk features without compromising utility from the risk-utility frontier. Different feature selection criteria can be applied based on the organizational setting and the priority of data privacy (or utility). Two feature selection criteria are proposed, namely *risk only* and *weighted cost*, to reflect individual and organizational privacy preferences. The feature with the largest value based on the chosen criterion will be selected for masking.

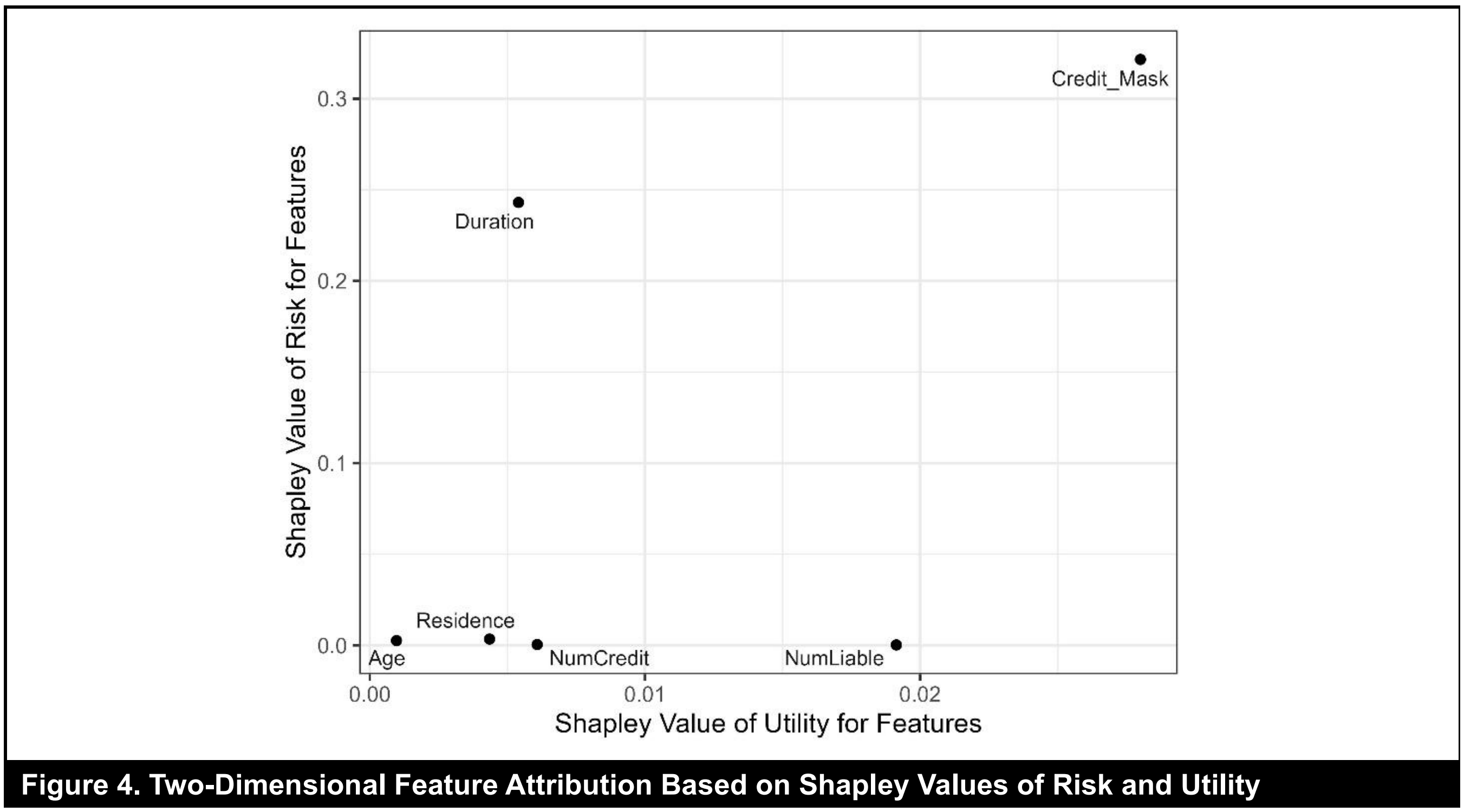

**Figure 4. Two-Dimensional Feature Attribution Based on Shapley Values of Risk and Utility**

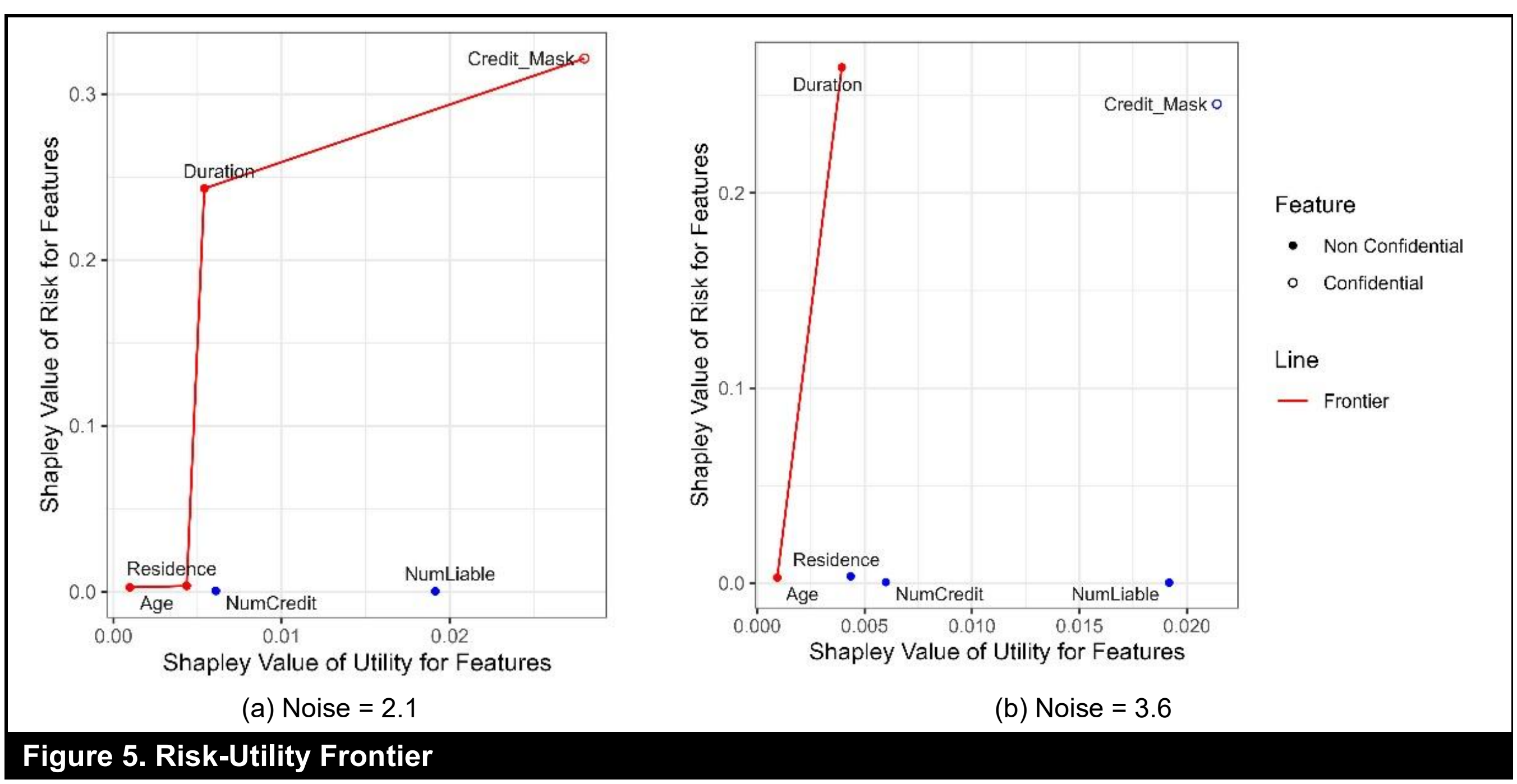

**Figure 5. Risk-Utility Frontier**

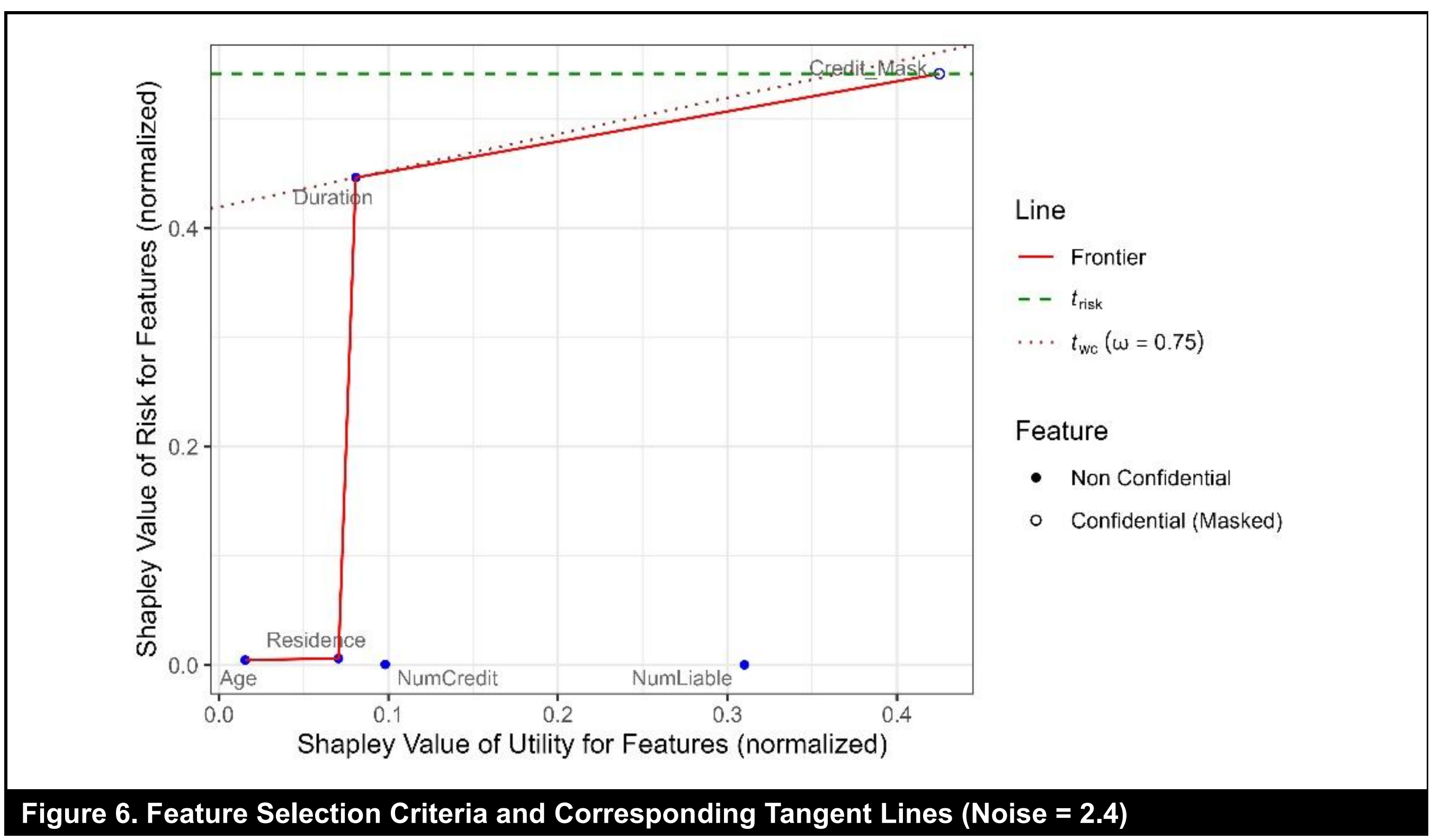

**Figure 6. Feature Selection Criteria and Corresponding Tangent Lines (Noise = 2.4)**

When the primary concern of an organization is data privacy, the risk-only selection criterion, i.e., $t_{risk} = \phi_i(r)$, should be used. When an organization has dual goals of risk reduction and utility preservation, a good selection criterion is the weighted cost, i.e., $t_{wc} = w\phi_i'(r) + (1-w)\big(-\phi_i'(u)\big)$, where $\phi_i'(\cdot) = \frac{\phi_i(\cdot)}{\sum \phi_i(\cdot)}$ are the normalized Shapley values. In this context, there are two types of costs for the organization—the cost of potential disclosure risk and the cost of potential utility loss. Since the utility Shapley value is a reverse measure of utility loss, we negate the utility Shapley value in the formula for $t_{wc}$. Additionally, we normalize Shapley values for risk and utility to make them unitless and on the same scale. The weight parameter $w$ in $t_{wc}$ represents the relative importance of privacy for the organization. Privacy preferences and benefits derived from improved data utility vary drastically among individuals, organizations, and even from one dataset to another (Xu & Zhang, 2022). Under different contexts, the weight ($w$) should be adjusted based on the priority of data privacy over data utility and can be selected by domain experts based on their individual and organizational privacy preferences. The two feature selection criteria work as two tangent lines that select the next feature to mask on the risk-utility frontier. In the illustrative example (Figure 6), the risk-only tangent curve suggests masking the confidential feature. However, the weighted-cost criterion selects the *Duration* feature to mask next.

Following the proposed feature selection procedure mentioned above, a feature from the candidate feature set (including both confidential and nonconfidential features) will be selected to be masked (steps 8-9 in Algorithm 1 below).[7] After masking the selected feature, the total disclosure risk and data utility can be reevaluated. The procedure in the proposed framework (Figure 3) should be executed iteratively until a desired level of data privacy (or disclosure risk) is reached.

## Experimental Studies

In this section, we evaluate the effectiveness and robustness of the proposed framework in protecting data privacy while maintaining data utility and compare it with benchmark methods. We first provide a comprehensive overview of the experiment setup, followed by an assessment in a simulated scenario, and conclude with evaluations using three real-world datasets.

[7] When certain features are intentionally shared or are crucial for organizational decisions, the proposed framework allows features to be excluded from masking and noise inserted to be limited to preserve essential information.

**Algorithm 1. Disclosure Risk Attribution and Data Masking Framework**

**Input:** Original dataset ($\mathcal{D}$), the confidential feature ($y$), and the feature to be predicted ($z$), incremental noise levels ($\{d, 2d, \ldots, kd\}$), feature selection criteria ($t$);
**Output:** Masked dataset ($\mathcal{D}'$), total disclosure risk, total data utility
1: Choose the data masking method $f$ and initialize the noise added $\Delta = \{\delta_1, \delta_2, \ldots, \delta_p\} = \mathbf{0}$;
2: Determine the privacy inferential model $g$ and data utility model $h$;
3: Decide characteristic functions $r(\hat{y}, y)$ and $u(\hat{z}, z)$ for disclosure risk and data utility respectively;
4: Update noise level for the confidential feature $\delta_p = \delta_p + d$;
5: Mask the confidential feature: $y' = f(y, \delta_p)$;
6: Infer the confidential feature $y$ from $\mathcal{D}'$ using $\hat{y} = g(x_1, x_2, \ldots, y')$ and predict $z$ using $\hat{z} = h(x_1, x_2, \ldots, y')$;
7: Compute disclosure risk $r(\hat{y}, y)$ and data utility $u(\hat{z}, z)$;
8: Calculate Shapley values $\phi_i(r)$ and $\phi_i(u)$ for each feature $x_i$, for $i \in \{1,2, \ldots, p\}$ with $x_p \equiv y'$;
9: Generate the risk-utility frontier of features $\mathcal{F} = \left\{x_i : \forall j \neq i, \left(\phi_i(r) > \phi_j(r)\right) \vee \left(\phi_i(u) < \phi_j(u)\right)\right\}$;
10: Select the next feature to mask following the selection criterion $t$: $x_i^* = \underset{x_i \in \mathcal{F}}{\operatorname{argmax}}\, t\left(\phi_i(r), \phi_i(u)\right)$;
11: **while** $\sum \delta_i \leq kd$ **do**
12: Update noise level for the selected feature $\delta_{i^*} = \delta_{i^*} + d$;
13: Mask selected feature $x_{i^*}$, $x'_{i^*} = f(x_{i^*}, \delta_{i^*})$;
14: Infer $y$ from $\mathcal{D}'$ using $\hat{y} = g(x_1, \ldots, x'_i, \ldots, y')$ and predict $z$ using $\hat{z} = h(x_1, \ldots, x'_i, \ldots, y')$;
15: Compute disclosure risk $r(\hat{y}, y)$ and data utility $u(\hat{z}, z)$;
16: Calculate Shapley values $\phi_i(r)$ and $\phi_i(u)$;
17: Update the risk-utility frontier of features $\mathcal{F}$;
18: Determine the next feature to mask: $x_i^* = \underset{x_i \in \mathcal{F}}{\operatorname{argmax}}\, t\left(\phi_i(r), \phi_i(u)\right)$;
19: **end while**

## *Experiment Setup*

We implement common data masking methods with different inferential and predictive models under a variety of risk and utility metrics, in combination with the two proposed feature selection criteria. The comprehensive experiment setups are summarized in Table 3.

**Data masking methods:** In the data masking procedure, we consider two types of data masking methods: additive perturbation and multiplicative perturbation. A random noise $e_j$ is generated following certain distributions and incorporated into the original feature based on additive (i.e., $y_j' = y_j + e_j$) or multiplicative (i.e., $y_j' = y_j \times e_j$) operations. Additive noise includes the bias-corrected additive data perturbation (Muralidhar et al., 1999) and the Laplace mechanism (Fernandes et al., 2021). Multiplicative noise includes the multiplicative noise scheme (Kim & Winkler, 2003) and the twin uniform distribution (Brackenbury et al., 2020).

**Inferential models and data utility applications:** At the predictive application and inferential disclosure stages, data analysts and potential intruders may apply different analytics methods (i.e., $g(\cdot)$ and $h(\cdot)$ functions). We consider two common analytics methods in the experiment: the linear model and regression trees.

**Disclosure risk and data utility metrics:** Under inferential disclosure, the confidential feature can be inferred using the released microdata. Distance-based metrics, such as absolute and Euclidean distance, are commonly used to assess disclosure risk by quantifying the gap between true and predicted confidential values (Li & Sarkar, 2011, 2014). The proportion of variability in the confidential feature that could be explained by other features in the shared microdata is another type of risk metric (e.g., $R^2$: see Muralidhar & Sarathy, 2006; Qian & Xie, 2015).

The loss of data utility after data masking can be measured from statistical characteristics of a single feature, including biases in estimating mean, standard deviation, marginal distribution of certain features (Li & Sarkar, 2006, 2014), or the relationships among features, such as product moment and rank-order correlations (Sarathy & Muralidhar, 2002). Furthermore, when the dataset is applied for general analytics purposes (e.g., classification), measures specific to these applications (e.g., precision, recall) can also be applied to evaluate the change in data utility.

| Table 3. Summary of Experiment Design | |
|---|---|
| **Experiment component** | **Experiment setup** |
| Data masking methods | • Additive methods<br>o Bias-corrected additive data perturbation (BCADP)<br>o Noise generated from the Laplace distribution (Laplace)<br>o Noise levels range from $\delta$ (0.3) to $k\delta$ (3) times the standard deviation of the feature to be masked<br>• Multiplicative methods<br>o Multiplicative noise scheme (MNS)<br>o Twin uniform distribution (TwinUnif)<br>o Noise levels range from $\delta$ (0.15) to $k\delta$ (1.5) times the standard deviation of the feature to be masked |
| Inferential models and data utility applications | • Linear regression (LM)<br>• Regression trees (CART) |
| Disclosure risk and data utility metrics | • $R^2$<br>• Average absolute distance (AAD)<br>• Root average squared distance (RASD) |
| Feature selection criteria | • Risk only ($t_{risk}$)<br>• Weighted cost ($t_{wc}$)<br>o Weight $w$ varies from 0.7 to 0.95 in increments of 0.05 |

Following the literature, we apply three common metrics, i.e., $R^2$, AAD, and RASD, to measure the statistical power of the released dataset to infer or predict the confidential feature or other information of interest. The $R^2$ metric can be interpreted as the fraction of variance of the dependent feature that has been accounted for by the other features in the released dataset, which can be defined as $R^2_{Risk} = 1 - \frac{\sum_j (y_j - \hat{y}_j)^2}{\sum_j (y_j - \bar{y}_j)^2}$ and $R^2_{Utility} = 1 - \frac{\sum_j (z_j - \hat{z}_j)^2}{\sum_j (z_j - \bar{z}_j)^2}$ for disclosure risk and data utility.[8] The distance-based metrics (i.e., AAD and RASD) denote the accuracy of predicted values compared to true values of the feature, which can be calculated as $AAD_{Risk} = \frac{1}{N}\sum_{j=1}^{N} |y_j - \hat{y}_j|$, $AAD_{Utility} = \frac{1}{N}\sum_{j=1}^{N} |z_j - \hat{z}_j|$, $RASD_{Risk} = \sqrt{\frac{1}{N}\sum_{j=1}^{N} (y_j - \hat{y}_j)^2}$, and $RASD_{Utility} = \sqrt{\frac{1}{N}\sum_{j=1}^{N} (z_j - \hat{z}_j)^2}$.

The metrics defined above evaluate the information disclosure risk and data utility of the entire shared dataset. The Shapley value approach assumes that the evaluation metric is monotonic with respect to the number of features in the dataset. That is, adding more features would not reduce the data utility. Therefore, the selection of the evaluation metric should align with the applied analytic model. It is important to note that when the evaluation metric is consistent with the objective function of the statistical or analytical model, the monotonicity assumption is satisfied. For example, in regression models, $R^2$ and ASD are based on squared errors, and both OLS and MLE models aim to minimize these squared errors. This makes these metrics monotonic as more features are added to the model. Additionally, Somanchi et al. (2022) demonstrated that, in decision tree models, the AUC is monotonic when more features are added.

Following the previous section, we can calculate the Shapley value of each feature based on its contribution to inferential disclosure risk, $\phi_i(r)$, and data utility, $\phi_i(u)$. The calculated Shapley values of features can help determine which feature to mask next.

**Feature selection criteria:** Following the literature, the first benchmark method (Benchmark I) implemented in this study only selects the confidential feature to mask. In addition, we implement a second benchmark method proposed by Li and Sarkar (2014), which randomly selects half of the nonconfidential features for data masking (Benchmark II).

For comparison, we evaluate the performance of our proposed method with two different feature selection criteria, i.e., risk only and weighted cost. The risk-only criterion selects a feature for masking based on each feature's contribution to disclosure risk using the Shapley value of risk, $\phi_i(r)$. The weighted-cost criterion incorporates the risk-utility trade-off in the feature selection. In our experiments, we chose various weight parameters (e.g., $w = 0.7, 0.75, \ldots, 0.95$) to reflect the relatively high priority on data privacy than data utility. In practice, privacy managers may adjust the weight parameter $w$ to reflect their preferences for the risk-utility trade-off.

[8] Some statistical/machine learning models may not produce an $R^2$ or may produce a negative $R^2$. In such scenarios, pseudo-$R^2$ can be a good substitute.

## Simulation Analysis

To demonstrate a complete application of the proposed framework and its computing time complexity, we conducted a simulation study on the general steps an organization follows from data collection to data sharing. Consider a bank that offers financial services and collects data from customers. The bank uses features from the microdata to classify customers as good or bad credit risks for future credit approvals and to predict the loan installment amounts customers can afford. As the bank compiles consumer information and constructs the microdata, it must adhere to data privacy regulations pertinent to consumer credit evaluation, which are discussed in the Problem Description and An Illustrative Example section. To illustrate the proposed method, we simulated microdata collected by a bank on its customers, using the Home Credit dataset from Kaggle[9] as a basis. In the simulation, categorical features were employed to create a tree-like partition of the dataset. Observations for the numerical features within each partition were generated based on statistical distributions estimated from the real dataset, including normal, lognormal, Poisson, zero-inflated Poisson, and mixture distributions. The initial simulated dataset included features on customers' identity (ID number, date of birth, zip code), demographics (age, gender, education level, etc.), and financial information (debt amount, number of credit lines, overdue debt amount, etc.). Identifiers and quasi-identifiers were removed from the data in compliance with regulations for identity protection (Kayaalp, 2017). The final simulated dataset had 30,000 observations and 33 features with no missing values.

Post de-identification, it is crucial to properly mask the confidential feature (i.e., in our case, credit amount) to mitigate potential disclosure risks, as recommended by the literature. Unlike the traditional methods that typically mask only the confidential feature, this study proposes a Shapley value-based feature attribution framework for microdata protection that considers both confidential and nonconfidential features due to the correlation that may exist in the dataset.

Results in Table 4 illustrate the performance of traditional methods (benchmark and random selection) and our proposed framework (risk only and weighted cost) based on the simulated data. The last two columns of the table demonstrate that, in most cases, the two proposed methods reduce risk more effectively than traditional approaches, while preserving data utility.

**Table 4. Experiment Results Using the Simulated Dataset**

| | Benchmark | | Random selection | | Risk only | | Weighted cost | | Comparison with benchmark | | |
|---|---|---|---|---|---|---|---|---|---|---|---|
| **Model** | **%δ$r$** | **%δ$u$** | **%δ$r$** | **%δ$u$** | **%δ$r$** | **%δ$u$** | **%δ$r$** | **%δ$u$** | **Random selection** | **Risk only** | **Weighted cost** |
| ***Masking method: BCADP*** | | | | | | | | | | | |
| LM | 16.538 | -0.003 | -4.536 | 1.334 | 52.638 | 3.84 | 37.85 | 0.01 | (**-21.07** & **-1.34**) | (**36.10** & **-3.85**) | (**21.31** & -0.01) |
| CART | 22.700 | 0.000 | -7.014 | 4.530 | 25.998 | 0.83 | 25.71 | 0.00 | (**-29.71** & **-4.53**) | (**3.30** & -0.83) | (3.01 & 0.00) |
| ***Masking method: Laplace*** | | | | | | | | | | | |
| LM | 23.066 | -0.004 | 5.710 | 0.886 | 55.171 | 1.74 | 61.19 | 1.30 | (-**17.36** & -0.89) | (**32.11** & **-1.75**) | (**38.12** & **-1.30**) |
| CART | 29.168 | 0.000 | 4.755 | 2.516 | 44.953 | 0.46 | 44.97 | 0.00 | (**-24.41** & **-2.52**) | (15.79 & -0.46) | (**15.80** & 0.00) |
| ***Masking method: MNS2*** | | | | | | | | | | | |
| LM | 4.102 | -0.001 | 0.230 | 16.005 | 4.102 | 0.00 | 4.10 | 0.00 | (**-3.87** & **-16.01**) | (0.00 & 0.00) | (0.00 & 0.00) |
| CART | 3.491 | 0.000 | 0.322 | 11.755 | 3.491 | 0.00 | 3.49 | 0.00 | (**-3.17** & -**11.76**) | (0.00 & 0.00) | (0.00 & 0.00) |
| ***Masking method: TwinUnif*** | | | | | | | | | | | |
| LM | 22.277 | -0.004 | 5.298 | 0.227 | 31.462 | 0.00 | 31.14 | 0.00 | (**-16.98** & -0.23) | (**9.19** & -0.01) | (**8.86** & -0.01) |
| CART | 23.468 | 0.000 | -7.322 | -0.118 | 23.706 | 0.00 | 23.71 | 0.00 | (**-30.79** & 0.12) | (0.24 & 0.00) | (0.24 & 0.00) |

**Note:** (1) The %$\Delta r$ and %$\Delta u$ columns denote the percentage of risk reduction or utility loss by the data masking method. (2) Values in parentheses in the last three columns represent the additional risk reduction and the additional utility loss by the specific data masking method compared to the benchmark. (3) Any differences in risk or utility reductions over 1% compared to the benchmark are marked in bold font, and values with underlines denote inferior performances compared to the benchmark.

[9] https://www.kaggle.com/competitions/home-credit-credit-risk-model-stability

In addition, we conducted a time complexity analysis by varying the number of observations ($n \in \{10000, 20000, 30000\}$) and the number of features ($p \in \{15,20,25\}$) in the microdata. For each combination, we replicated the masking procedure 100 times in parallel on a Linux server cluster using R version 4.3.1, with each task requesting 4GB RAM and 1 CPU core (~ 2.3GHz). The average computing time of the analyses is illustrated in Appendix Table B1. Based on this back-of-the-envelope analysis, the computational cost increases linearly with the number of records but exponentially with the number of features in the dataset. The computing time can be further reduced using the high-performance computing resources available to industry practitioners.

## Description of Three Real-World Datasets

To evaluate the performance of the proposed framework, the experiments were designed and conducted on three real-world datasets. Given the focus on quantitative features of this paper, we selected quantitative features from the datasets for our experiments.

### Credit History

The first dataset was obtained from the UCI Machine Learning Repository (Hofmann, 1994).[10] This dataset classifies consumers described by a set of features as good or bad credit risks. It has 1,000 customer records with 20 features (7 quantitative and 13 categorical) used by a bank to evaluate credit applications. Some quantitative features are demographic or economic in nature, and other features are banking or credit related. Details of the features are described in Appendix Table A1. In line with prior studies (e.g., Li & Sarkar, 2014), the feature *CreditAmount* was considered to be the confidential feature, while the feature *Installment_Rate* was used as a utility feature (i.e., feature to be predicted).

### MIS Faculty Salary Offers

The Association for Information Systems conducts annual surveys of MIS faculty salary offers (Galletta, 2004). This salary dataset consists of 832 applicants who received offers and reported complete information in the survey from 2003 to 2016. Details of the features in this salary dataset are described in Appendix Table A2. Following prior studies (Li & Sarkar, 2011), the offered *Salary* was considered to be the confidential feature and the annual discretionary research budget (*Budget*) was used as the utility feature.

### CRSP/Compustat Dataset

The third dataset was collected from the CRSP/Compustat Merged (CCM) database. Duplicated records for the same firms were removed, categorical variables were omitted, and features with high correlations (over 0.9) were eliminated from the dataset. To exclude extreme observations from the dataset, we applied 96% winsorization that set extremely large (small) values of the features to 98th (or 2nd) percentile values. The final sample includes 3,704 firms' annual financial records for the year 2021, with 47 quantitative features. The annual closing price of the firms' stock (*PRCC_C*) was used as the utility variable, as stock prices are publicly observable and predicting stock prices is a common practice. We used the comprehensive income (*CI*) as the confidential feature to be protected because the comprehensive income is unknown to the market before a company releases its financial statement. Descriptions of the dataset can be found at CRSP.[11]

## Experimental Results

This subsection presents the experimental results for the benchmark methods and our proposed framework. For illustration, we focus on the credit and salary datasets to demonstrate the differences between the traditional method of masking only confidential features and the proposed data masking framework in the next two subsections. Then, the following subsection comprehensively compares all data masking methods under a spectrum of application scenarios, including the large-scale CRSP/Compustat dataset. The final subsection examines the application of the proposed data masking framework to scenarios where multiple confidential features exist.

### Performance of the Benchmark Method

We first examine the performance of the benchmark method, which focuses on masking only the confidential feature(s) for data privacy protection. As the confidential feature is masked, the disclosure risk can be reduced, but the utility of the dataset will also decrease. The trade-off between disclosure risk and data utility can be demonstrated in a risk-utility map, in which the data utility is illustrated on the $x$-axis, while the potential disclosure risk is represented on the $y$-axis. A method that can reduce the disclosure risk while maintaining data utility is preferred.

[10] https://archive.ics.uci.edu/ml/datasets/statlog+(german+credit+data)

[11] https://www.crsp.org/research/crsp-compustat-merged-database/

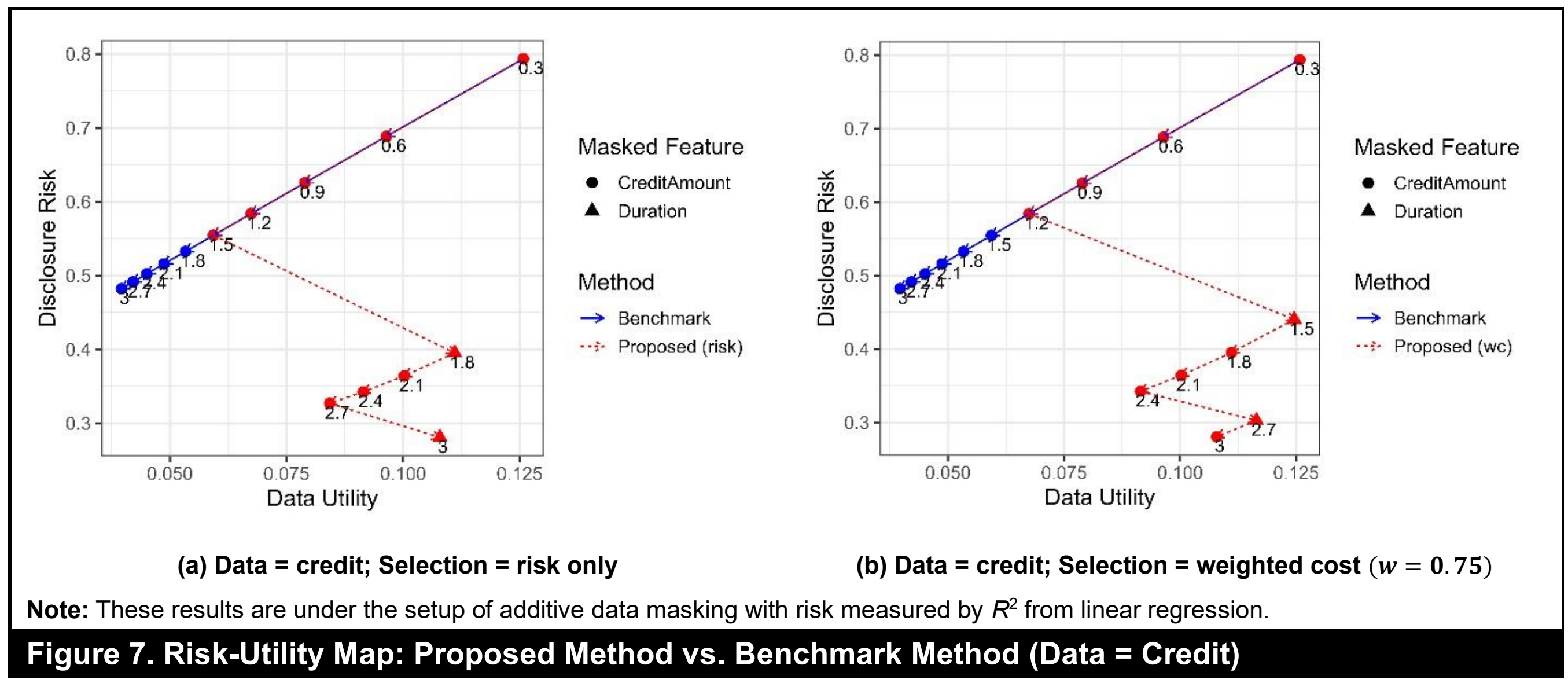

**(a) Data = credit; Selection = risk only** **(b) Data = credit; Selection = weighted cost $(w = 0.75)$**

**Note:** These results are under the setup of additive data masking with risk measured by $R^2$ from linear regression.

**Figure 7. Risk-Utility Map: Proposed Method vs. Benchmark Method (Data = Credit)**

**(a) Data = salary; Selection = risk only** **(b) Data = salary; Selection = weighted cost $(w = 0.75)$**

**Note:** These results are under the setup of additive data masking with risk measured by $R^2$ from linear regression.

**Figure 8. Risk-Utility Map: Proposed Method vs. Benchmark Method (Data = Salary)**

The performance of the benchmark method, which only masks the confidential feature in a dataset, is depicted as the solid line in Figures 7 and 8. The $y$-axis represents the disclosure risk of a released dataset, which is measured by the $R^2$ of a simple linear inferential model. A greater $R^2$ value means that the true value of the confidential feature can be largely explained by the released dataset. The utility of a dataset in the $x$-axis is also measured using the $R^2$ of a linear predictive model, representing the utility of the dataset to predict a utility feature.

The value labeled in Figures 7 and 8 denotes the level of noise added to the dataset. Following the data masking procedure, as more noise is added to the confidential feature, the disclosure risk of the released dataset decreases. Meanwhile, the utility of the dataset also diminishes. More importantly, as the masking approach proceeds, the marginal decrease in the total disclosure risk level diminishes as more noise is added to the feature. In other words, once the noise level gets to a certain point, masking the confidential feature is no longer effective for reducing the disclosure risk, but it may lead to a significant reduction in data utility. Therefore, instead of masking only the confidential feature, other nonconfidential features should also be considered if masking them could more effectively reduce the disclosure risk while maintaining the data utility.

## Performance of the Proposed Method

To evaluate the performance of the proposed framework in Figure 3, we compare its performance with the benchmark method. In Figures 7 and 8, the dashed curve represents the performance of the proposed methods in a risk-utility map, while the benchmark method is denoted with a solid line. In general, the proposed method outperforms (is at least as well as) the benchmark approach in reducing the disclosure risk, while still preserving the utility of a dataset.

As is evidenced by the risk-utility frontier (Figure 5), masking the confidential feature may not be the most efficient solution. The Shapley value-based feature attribution approach evaluates how each feature contributes to the disclosure risk and data utility and selects features to mask. In the running credit example (Figure 7), the proposed framework can effectively select another nonconfidential feature, *Duration*, to mask, leading to a further decrease of risk. In the salary example (Figure 8), the proposed framework selects two nonconfidential features, *Summer* and *Move_Supp*, to mask, which further reduces the risk level of the dataset while maintaining its utility.

The Shapley value-based feature attribution approach extends the risk-utility trade-off to the feature level. When selecting the next feature to mask, different priorities assigned to risk or utility might lead to different results. A comparison of Figure 8(a) and 8(b) shows that by following different feature selection criteria (e.g., risk only or weighted cost), the proposed framework results in different levels of risk and utility of the released dataset. In general, the proposed method performs better in terms of disclosure risk protection and data utility preservation compared to the benchmark approach, regardless of the feature selection criteria.

It is worth noting that although the proposed method can suggest other nonconfidential features following the attribution in risk and utility, it does not always suggest other features to mask. When the masked confidential feature (e.g., *Salary*) is the suggested feature, the proposed method would select the confidential feature and perform as well as the benchmark method.

## Comprehensive Comparison Among Data Masking Methods

For a comprehensive evaluation, we compared the proposed method to benchmark methods in terms of the effectiveness of risk reduction and utility preservation under all combinations of the setup listed in Table 3. For each setup, we ran 1,000 iterations and took the average result to control for the randomness in the data masking process. The main results are reported in Table 5(a, b, c), and the computational time costs are documented in Appendix Table B2.

The first three columns of Table 5, respectively, represent the masking technique, the inferential/utility model, and the evaluation metrics used. In the table, the values in the parentheses, respectively, denote the percentage reduction in risk ($\%\Delta r$) and utility ($\%\Delta u$) of the specific data masking method. Greater values in risk reduction (i.e., the first value in the parentheses) indicate better performance of the data masking method, and smaller values in utility loss (i.e., the second value in the parentheses) signal better performance in preserving data utility. Columns 4-7 illustrate the performance of different data masking methods under the credit (Table 5a), salary (Table 5b), and Compustat (Table 5c) datasets, respectively. To highlight their relative performance, we compared the data masking methods to Benchmark I, which masks only the confidential feature, and marked any differences in risk reduction or utility loss greater than 1% with bold font. For any method that performed worse than Benchmark I, the value of risk reduction or utility loss is marked with underlining.

As shown in Table 5, data masking generally leads to a decrease in disclosure risk for both the benchmark and the proposed methods (i.e., risk only and weighted cost). Meanwhile, the utility of the dataset also decreases in most cases. Data utility actually increases after masking under a few settings. This is potentially due to the choice of the feature for the prediction task, and the utility increases are marginal. Comparisons between the benchmark methods (the first two columns under each use case) and the proposed methods (the last two columns under each use case) demonstrate that the random feature selection for masking is not an effective strategy for privacy protection, as it cannot contribute to further risk reduction compared to masking the confidential feature only. The risk-only and weighted-cost approaches under the proposed framework outperform (or at least perform as well as) the benchmark methods in terms of risk reduction in all cases. Meanwhile, the proposed framework also leads to less utility loss than the benchmarks in most cases. In only a few cases (e.g., protecting the salary dataset with BCADP using linear models measured with AAD), the risk-only and weighted-cost approaches lead to slightly more utility loss. But compared to the significant reduction in risk, the utility losses are marginal.

## Application to Scenarios with Multiple Confidential Features

In practice, multiple confidential features may exist in the microdata. When the dataset is released or shared, all the confidential features then become subject to potential inferential attacks by intruders. Disclosure of any of the confidential features can lead to privacy losses. Hence, it is important to account for all confidential features, comprehensively evaluate the inferential disclosure risk of the released microdata, and fairly measure the importance of each feature to the aggregated disclosure risk. In this subsection, we examine the application of the proposed data masking framework to scenarios with multiple confidential features.

**Table 5a. Experiment Results: Comparison of Data Masking Methods (Data: Credit)**

| Mask | Model | Metric | Benchmark I: Confidential only | Benchmark II: Random selection | Risk only | Weighted cost |
|---|---|---|---|---|---|---|
| BCADP | LM | $R^2$ | (38.86, 66.45) | (**13.90**, **-11.02**) | (**63.38**, **18.72**) | (**63.41**, **18.48**) |
| | | AAD | (43.78, 80.34) | (**22.78**, **-12.28**) | (**73.89**, **3.63**) | (**73.91**, **3.85**) |
| | | RASD | (48.40, 67.21) | (**19.64**, **-11.64**) | (**71.20**, **19.27**) | (**71.22**, **19.09**) |
| | CART | $R^2$ | (35.44, 52.75) | (**12.00**, **-20.78**) | (**47.08**, **15.02**) | (**43.32**, **15.15**) |
| | | AAD | (39.97, 61.20) | (**17.55**, **-26.42**) | (**55.99**, **3.77**) | (**53.79**, **3.69**) |
| | | RASD | (45.60, 53.87) | (**17.93**, **-22.34**) | (**57.03**, **15.60**) | (**53.51**, **15.09**) |
| Laplace | LM | $R^2$ | (52.22, 85.01) | (**31.55**, **12.84**) | (**87.41**, **26.56**) | (**87.41**, **26.56**) |
| | | AAD | (57.31, 96.76) | (**48.57**, **16.69**) | (**90.57**, **20.37**) | (**90.42**, **20.80**) |
| | | RASD | (62.91, 85.50) | (**42.19**, **13.10**) | (**91.11**, **27.36**) | (**91.11**, **27.36**) |
| | CART | $R^2$ | (49.78, 71.43) | (**25.70**, **10.93**) | (**72.44**, **26.04**) | (**72.36**, **26.70**) |
| | | AAD | (54.83, 76.47) | (**35.64**, **11.81**) | (**78.26**, **26.25**) | (**78.19**, **26.07**) |
| | | RASD | (60.96, 72.50) | (**36.42**, **11.22**) | (**79.93**, **27.04**) | (**79.86**, **27.61**) |
| MNS | LM | $R^2$ | (44.46, 73.61) | (**32.26**, **27.20**) | (**65.77**, **48.91**) | (**66.12**, **48.26**) |
| | | AAD | (51.67, 84.78) | (**39.48**, **33.28**) | (**70.30**, **58.40**) | (**70.66**, **58.44**) |
| | | RASD | (55.04, 74.32) | (**42.31**, **27.87**) | (**73.95**, **49.84**) | (**74.23**, **49.42**) |
| | CART | $R^2$ | (38.76, 56.32) | (**14.36**, **-0.01**) | (**51.67**, **27.93**) | (**51.10**, **27.40**) |
| | | AAD | (45.64, 63.89) | (**20.95**, **-0.25**) | (**58.25**, **31.93**) | (**56.18**, **31.06**) |
| | | RASD | (49.27, 57.97) | (**20.97**, **-0.19**) | (**61.55**, **29.30**) | (**61.08**, **29.56**) |
| TwinUnif | LM | $R^2$ | (42.73, 63.96) | (**13.15**, **12.61**) | (42.98, **62.81**) | (**44.08**, **62.75**) |
| | | AAD | (56.03, 72.42) | (**28.31**, **15.18**) | (56.10, **72.03**) | (**57.69**, **72.02**) |
| | | RASD | (62.17, 65.09) | (**30.76**, **13.16**) | (62.36, **63.95**) | (**63.18**, **63.78**) |
| | CART | $R^2$ | (32.82, 54.51) | (**9.41**, **6.18**) | (32.72,53.86) | (32.73, **53.38**) |
| | | AAD | (42.89, 63.25) | (**20.18**, **10.68**) | (42.87,63.21) | (42.88, 62.95) |
| | | RASD | (50.49,56.59) | (**21.08**, **6.67**) | (50.39,55.94) | (50.40, **55.46**) |

**Note:** (1) Values in the parentheses are the percentages of risk reductions and utility losses by specific approach, i.e., ($\%\Delta r_t$, $\%\Delta u_t$). Higher values of risk reductions and lower values of utility losses indicate better performance of the selected approach as compared to the traditional method (i.e., masking the confidential feature only). (2) In this table, for the proposed method with the weighted-cost selection criterion, we report the results based on $w = 0.75$. The experiment results for other weight parameter values are qualitatively the same. (3) Any differences in risk or utility reductions over 1% compared to Benchmark I, masking only the confidential feature, are marked in bold font, and underlined values denote inferior performance compared to Benchmark I.

**Table 5b. Experiment Results: Comparison of Data Masking Methods (Data: Salary)**

| Mask | Model | Metric | Benchmark I: Confidential only | Benchmark II: Random selection | Risk only | Weighted cost |
|---|---|---|---|---|---|---|
| BCADP | LM | $R^2$ | (24.95, -0.19) | (**2.82**, **11.45**) | (**33.65**, **1.05**) | (**33.37**, 0.24) |
| | | AAD | (34.15, 1.02) | (**2.89**, **19.75**) | (**47.39**, **4.34**) | (**46.75**, **2.31**) |
| | | RASD | (34.57, -0.22) | (**4.62**, **12.47**) | (**43.72**, **1.19**) | (**43.47**, 0.28) |
| | CART | $R^2$ | (17.86, 3.90) | (**7.75**, **-12.77**) | (**27.97**, **-3.22**) | (**28.35**, **-5.01**) |
| | | AAD | (22.11, 6.33) | (**9.74**, **-18.30**) | (**31.24**, **-1.73**) | (**26.86**, **-2.66**) |
| | | RASD | (25.78, 4.32) | (**12.02**, **-14.56**) | (**37.75**, **-4.27**) | (**37.97**, -6.28) |
| Laplace | LM | $R^2$ | (34.12, -0.27) | (**16.34**, **10.72**) | (**53.21**, **4.38**) | (**51.62**, **1.98**) |
| | | AAD | (47.25, 1.11) | (**25.06**, **19.81**) | (**67.86**, **9.32**) | (**67.60**, **6.24**) |
| | | RASD | (46.46, -0.31) | (**25.62**, **11.70**) | (**64.49**, **4.99**) | (**63.09**, **2.28**) |
| | CART | $R^2$ | (20.94, 6.05) | (**17.07**, **-10.04**) | (**39.86**, **-5.59**) | (**39.40**, **-8.25**) |
| | | AAD | (29.15, 9.26) | (**23.25**, **-13.45**) | (**44.22**, **-3.68**) | (**41.56**, **-7.27**) |
| | | RASD | (30.95, 6.71) | (**25.80**, **-11.44**) | (**51.42**, **-6.55**) | (**50.71**, **-9.06**) |
| MNS | LM | $R^2$ | (26.87, -0.26) | (**6.48**, **11.94**) | (**28.41**, -0.21) | (**28.51**, -0.24) |
| | | AAD | (38.87, 1.06) | (**11.48**, **16.96**) | (**41.52**, 1.17) | (**41.64**, 1.10) |
| | | RASD | (39.39, -0.30) | (**11.97**, **13.01**) | (**40.95**, -0.25) | (**41.06**, -0.28) |

| | CART | $R^2$ | (22.92, 3.69) | (**8.72**, **-11.06**) | (**25.99**, **0.45**) | (**26.42**, -**0.16**) |
|---|---|---|---|---|---|---|
| | | AAD | (31.28, 6.28) | (**13.83**, **-16.51**) | (**35.09**, **2.12**) | (**32.96**, 1.21) |
| | | RASD | (33.67, 4.06) | (**14.73**, **-12.62**) | (**37.29**, **0.41**) | (**37.73**, **-0.65**) |
| TwinUnif | LM | $R^2$ | (35.62, -0.48) | (**27.58**, **4.02**) | (**42.68**, -0.06) | (**42.72**, -0.19) |
| | | AAD | (46.60, 1.60) | (**37.79**, **9.19**) | (**55.31**, 2.18) | (**55.45**, 1.83) |
| | | RASD | (48.76, -0.55) | (**40.09**, **4.44**) | (**55.62**, -0.09) | (**55.64**, -0.22) |
| | CART | $R^2$ | (20.81,4.33) | (**18.85**, **-12.63**) | (21.15,3.48) | (21.36, **1.88**) |
| | | AAD | (25.87,7.63) | (**23.24**, **-14.83**) | (26.25,7.16) | (26.09, 5.51) |
| | | RASD | (31.13,4.80) | (**28.55**, **-14.37**) | (31.60, **3.64**) | (31.86, 1.61) |

**Note:** See Table 5a

**Table 5c. Experiment Results: Comparison of Data Masking Methods (Data: Compustat)**

| Mask | Model | Metric | Benchmark I: Confidential only | Benchmark II: Random selection | Risk only | Weighted cost |
|---|---|---|---|---|---|---|
| BCADP | LM | $R^2$ | (3.07, 0.04) | (**-4.33**, 0.51) | (**7.18**, 0.18) | (**6.12**, -0.12) |
| | | AAD | (-5.42, 0.08) | (**-17.74**, 0.91) | (**-3.38**, -0.24) | (-6.10, -0.20) |
| | | RASD | (6.37, 0.06) | (**-11.35**, 0.65) | (**15.69**, 0.40) | (**13.59**, -0.15) |
| | CART | $R^2$ | (8.33, 0.00) | (**1.73**, **4.26**) | (**9.72**, **-2.92**) | (8.80, **-1.52**) |
| | | AAD | (7.60, 0.00) | (**3.29**, **4.94**) | (**13.66**, **2.90**) | (**11.18**, **-3.57**) |
| | | RASD | (15.88, 0.00) | (**3.75**, **5.22**) | (**18.37**, **-1.40**) | (16.58, **-1.95**) |
| Laplace | LM | $R^2$ | (4.77, 0.07) | (**0.59**, 0.37) | (**10.02**, 0.50) | (**9.06**, -0.01) |
| | | AAD | (-3.65, 0.13) | (**-0.86**, 0.77) | (**7.22**, -0.11) | (**2.74**, -0.25) |
| | | RASD | (10.02, 0.09) | (**1.31**, 0.46) | (**21.64**, 0.74) | (**19.18**, 0.00) |
| | CART | $R^2$ | (12.09, 0.00) | (**2.80**, **2.61**) | (**15.16**, **3.90**) | (12.49, **-4.24**) |
| | | AAD | (11.13, 0.00) | (**4.92**, **2.38**) | (**19.50**, **4.78**) | (**15.48**, **-2.60**) |
| | | RASD | (22.33, 0.00) | (**6.25**, **3.20**) | (**27.64**, **5.51**) | (**24.38**, **-3.25**) |
| MNS | LM | $R^2$ | (5.44, 0.06) | (**-0.83**, **4.32**) | (**9.04**, 0.64) | (**8.82**, 0.42) |
| | | AAD | (11.34, 0.18) | (**-1.01**, **5.20**) | (**18.60**, 0.46) | (**18.60**, 0.44) |
| | | RASD | (11.80, 0.08) | (**-2.23**, **5.40**) | (**19.42**,**1.64**) | (**17.98**, 0.52) |
| | CART | $R^2$ | (9.64, 0.00) | (**1.80**, **3.40**) | (10.54, **-1.18**) | (10.21, -0.75) |
| | | AAD | (14.36, 0.00) | (**2.95**, **2.62**) | (**16.87**, **-1.84**) | (**16.74**, **-1.78**) |
| | | RASD | (18.19, 0.00) | (**4.02**, **4.16**) | (**19.41**, **-1.26**) | (18.96, -0.86) |
| TwinUnif | LM | $R^2$ | (9.87, 0.13) | (**0.67**, 0.09) | (10.62, 0.31) | (10.60, 0.28) |
| | | AAD | (21.86, 0.35) | (**3.02**, 0.13) | (**25.80**, 0.59) | (**25.82**, 0.58) |
| | | RASD | (26.39, 0.17) | (**3.71**, 0.12) | (**27.68**, 0.38) | (**27.66**, 0.36) |
| | CART | $R^2$ | (14.07,0.00) | (0.15, 0.19) | (13.84, -0.81) | (13.99, -0.25) |
| | | AAD | (19.10,0.00) | (**-0.49**, -0.48) | (19.25, **-1.38**) | (19.15, -0.68) |
| | | RASD | (28.12,0.00) | (**0.45**, 0.22) | (27.95, -0.52) | (28.06, -0.19) |

**Note:** See Table 5a

To apply the proposed data masking framework to scenarios with multiple confidential features, firms or organizations only need to modify the measure for inferential disclosure risk with a comprehensive and aggregated risk evaluation metric in the framework. In other words, in the framework illustrated in Figure 3, confidential feature $y$ is extended to $\boldsymbol{y} = \{y_1, \ldots, y_q\}$ and inferential disclosure risk metric $r(\hat{y}, y)$ is extended to $r(\hat{\boldsymbol{y}}, \boldsymbol{y})$. Then, based on the overall disclosure risk measure, the feature attribution steps in the proposed framework can effectively and fairly attribute the total disclosure risk and data utility to each individual feature. Next, the feature selection process can select important features for masking based on the firm's preferences for the risk-utility trade-off, followed by data masking to insert noise into the selected feature.

To evaluate this extended method, we used the salary dataset and applied our framework to protect two confidential features (i.e., *Salary* and *Summer*). $R^2$ was selected as the risk evaluation metric for each confidential feature. Thus, the overall risk level of the dataset can be calculated by the summation of the disclosure risk of each confidential feature (equally weighted in the experiment), i.e., $r(\hat{\boldsymbol{y}}, \boldsymbol{y}) = R^2_{salary|\mathcal{D}'} + R^2_{summer|\mathcal{D}'}$.

**Table 6. Experiment Results under Multiple Confidential Features (Data: Salary)**

| Mask | Model | Benchmark I ($BM_I$): Confidential only | | Benchmark II: Random selection | | Risk only | | Weighted cost | |
|---|---|---|---|---|---|---|---|---|---|
| | | $\%\Delta r$ | $\%\Delta u$ | $\%\Delta r$ | $\%\Delta u$ | $\%\Delta r$ | $\%\Delta u$ | $\%\Delta r$ | $\%\Delta u$ |
| BCADP | LM | 20.07 | -0.34 | 8.13<br>**(-11.94)** | 12.04<br>**(-12.38)** | 22.97<br>**(2.90)** | 0.44<br>(-0.78) | 20.81<br>(0.74) | -0.34<br>(0.00) |
| | CART | 29.63 | 0.37 | 17.08<br>**(-12.55)** | -8.10<br>**(8.47)** | 34.90<br>**(5.27)** | -1.83<br>**(2.20)** | 34.47<br>**(4.84)** | -3.97<br>**(4.34)** |
| Laplace | LM | 27.40 | -0.50 | 16.90<br>**(-10.50)** | 12.41<br>**(-12.91)** | 30.75<br>**(3.35)** | 0.51<br>**(-1.01)** | 28.65<br>**(1.25)** | -0.59<br>(0.09) |
| | CART | 36.27 | 3.19 | 28.96<br>**(-7.31)** | -5.08<br>**(8.27)** | 46.88<br>**(10.61)** | -1.82<br>**(5.01)** | 44.54<br>**(8.27)** | -6.41<br>**(9.60)** |
| MNS | LM | 29.07 | -1.37 | 17.07<br>**(-12.00)** | 13.81<br>**(-15.18)** | 32.73<br>**(3.66)** | 0.64<br>**(-2.01)** | 30.57<br>**(1.50)** | -0.77<br>(-0.60) |
| | CART | 29.68 | 0.80 | 16.59<br>**(-13.09)** | -7.48<br>**(8.28)** | 31.23<br>(**1.55)** | 0.21<br>(0.59) | 31.24<br>**(1.56)** | -1.81<br>**(2.61)** |
| TwinUnif | LM | 27.55 | -0.69 | 16.84<br>**(-10.71**) | 4.61<br>(-5.30) | 29.37<br>**(1.82)** | -1.13<br>(0.44) | 29.32<br>**(1.77)** | -1.18<br>(0.49) |
| | CART | 20.28 | 2.57 | 13.66<br>**(-6.62)** | -5.32<br>**(7.89)** | 26.62<br>**(6.34)** | 4.47<br>**(-1.90)** | 26.46<br>**(6.18)** | 3.46<br>(-0.89) |

**Note:** (1) $\%\Delta r_t$ and $\%\Delta u_t$ columns denote the percentage of risk reduction or utility loss by the data masking method. (2) Values in parentheses in the last three pairs of columns represent the additional risk reduction (i.e., $\%\Delta r_{t_{wc}} - \%\Delta r_{BM_I}$) and the additional utility loss (i.e., $\%\Delta u_{BM_I} - \%\Delta u_{t_{wc}}$), respectively, by the specific data masking method compared to Benchmark I, which masks the confidential feature only. (3) Any differences in risk or utility reductions over 1% compared to Benchmark I, which masks the confidential feature only, are marked in bold font, and underlined values denote inferior performances compared to Benchmark I.

The results in Table 6 show that the proposed approaches (i.e., risk only and weighted cost) can always effectively reduce the risk level. From the utility perspective, the risk-only selection criterion may lead to more utility decreases than the benchmark methods due to its focus on risk reduction. The weighted-cost selection criterion can effectively strike a balance between the risk-utility trade-off, providing additional protection to the data, while effectively preserving the utility of the dataset compared to the benchmark methods. Although the performance of the proposed framework does not depend on the number of confidential features, the computational time will increase linearly with the number of confidential features influenced by the risk evaluation metric.

## Future Work and Conclusion

Data privacy protection has become an important issue for consumers, companies, and policymakers. To address concerns about data privacy, the literature has proposed various data masking techniques to prevent the disclosure of confidential information. Although direct disclosure risks can be alleviated by data masking, potential intruders can use other features in a shared dataset (including the masked confidential feature and other nonconfidential features) to infer the confidential information. This study addresses this important issue and focuses on protecting data against inferential disclosure risks. This paper proposes a Shapley value-based feature attribution approach to quantify the contributions to disclosure risk and data utility from both confidential and nonconfidential features. We propose a data masking framework to select features for masking to protect the dataset at the feature level. Experimental results show that the proposed method can effectively reduce disclosure risks while preserving the utility of the dataset.

This study makes several contributions to the data privacy literature. To begin with, the proposed framework takes a holistic view of data masking that considers both confidential and nonconfidential features in data masking. Specifically, this paper is the first to utilize the Shapley value-based feature attribution approach to provide a fair and unique allocation of disclosure risk and data utility for both confidential and nonconfidential features in a dataset. Empirical evidence shows that nonconfidential features can lead to higher disclosure risks compared to a masked confidential feature. Secondly, the trade-off between risk and utility has been discussed at the dataset level in prior studies. Based on the two-dimensional Shapley values of disclosure risk and data utility evaluated by the proposed method, we are able to address the trade-off at the feature level by identifying candidate features to mask using the risk-utility frontier and selecting features to mask following different criteria. Lastly, the proposed data masking framework is model agnostic, meaning it can be applied in combination with a variety of data masking approaches, predictive/inferential models, and risk/utility metrics commonly applied in the literature. Experiments based on three real-world datasets (Table 5) consistently showed the advantage of the proposed framework under different settings.

This study has some significant practical implications. When protecting microdata from inferential disclosure risk, it is not sufficient to mask only confidential features. Firms/organizations should consider the relationship between confidential and nonconfidential features and adopt a holistic view when implementing data masking techniques. Following the framework proposed in this study, firms/organizations can assess the inferential disclosure risks of the data to be released and take actions to protect the confidential information. Before releasing a dataset, in addition to masking the confidential feature, organizations can apply the Shapley value-based feature attribution approach to identify features that contribute the most (least) to disclosure risk (data utility). Appropriate feature selection criteria should be chosen based on the priority of data utility and privacy concerns for different stakeholders. For example, since users and regulators usually put a high priority on data privacy, the risk-only criterion should be chosen. However, for internal and external data analysts, the weighted-cost criterion can strike a balance towards the risk-utility trade-off and maintain the utility of the data while meeting the privacy requirements. After that, the dataset should be masked and then reevaluated until a desired level of risk has been reached or a pre-specified level of noise has been allocated. The stopping strategies for the data masking procedure should be carefully selected according to the stakeholders' preferences for the risk-utility trade-off.

This work is not without limitations, which point to several promising future research directions. This study focuses on masking methods for quantitative features. Masking methods for categorical features could be implemented following the proposed framework in future extensions. Future work could also extend our framework to algorithm-based disclosure, the risk of disclosing details of data masking algorithms to potential adversaries (Jin et al., 2011). Another potential research direction would be to explore various options of a coalition-based Shapley value data masking approach that simultaneously determines the subset of features and its corresponding noise allocation schema for mitigating risk while maintaining utility. Additionally, masked microdata might be susceptible to data reconstruction attacks by adversaries (Garfinkel et al., 2019; Huang et al., 2005). Since this paper aims to develop an efficient and effective feature evaluation and selection mechanism in data masking, rather than to advance specific data masking methods, we suggest evaluating the performance of the proposed framework under potential reconstruction attacks as a future research direction. Furthermore, this study has important implications for future research on feature attribution beyond the data privacy protection setting. The proposed Shapley value-based data masking framework could lead to a new stream of research on evaluating risk-utility trade-off at the feature level and could be applied to broader contexts, such as the pricing of data features.

## Acknowledgments

The authors are grateful to the senior editor, Balaji Padmanabhan, and the Associate Editor for their valuable guidance, and to the three anonymous reviewers for their constructive feedback throughout the review process.

## About the Authors

**Xinxue (Shawn) Qu** is an assistant professor at the Mendoza College of Business, University of Notre Dame. His research focuses on developing analytical models and proposing system design solutions that guide business innovations and provide insights for decision makers. His work focuses on three main research areas: (1) innovation diffusion, (2) competitive strategies in innovation management, and (3) economics of artificial intelligence (AI) and machine learning. His research has been published in top business journals such as *MIS Quarterly* and *Production and Operations Management*, and *Decision Sciences*. ORCiD: 0000-0002-3316-2052

**Francis Bilson Darku** is an assistant professor of IT, analytics, and operations at the Mendoza College of Business, University of Notre Dame. His research interests include sequential analysis, nonparametric statistics and econometrics. He focuses on developing sequential methodologies under a distribution-free environment with applications in economics, business, psychology and other areas. Francis obtained a B.Sc. in actuarial science from Kwame Nkrumah University of Science and Technology, Kumasi–Ghana, and an M.Sc. and a Ph.D. in statistics from University of Texas at Dallas. ORCiD: 0000-0001-8639-8862

**Hong Guo** is a professor of information systems at the W. P. Carey School of Business, Arizona State University. Hong studies emerging IT phenomena by characterizing their key design features, examining firms' corresponding strategies, and analyzing the impacts of related IT policies. Her areas of expertise include digital platforms, GenAI business applications, business data visualization, etc. She teaches business data visualization in various graduate programs at W. P. Carey. ORCiD: 0000-0001-6028-8155

# Appendix A

## Data Description

**Table A1. Feature Description of Credit Data**

| Feature name | Description |
|---|---|
| *Age* | Age of the applicant in years |
| *Residence* | The length of residence at the current address (in years) |
| *Duration* | Duration in months |
| *NumCredit* | Number of existing credit lines at this bank |
| *NumLiable* | Number of people for whom the bank is liable for providing maintenance |
| *InstallRate* | Installment rate in percentage of disposable income |
| *CreditAmount* | Credit amount |

**Table A2. Feature Description of Salary Data**

| Feature Name | Description |
|---|---|
| *Salary* | Salary offered |
| *Experience* | Number of years of teaching experience |
| *TopJ* | Number of top journal publications |
| *RefJ* | Number of refereed journal publications |
| *OtherPubs* | Number of publications accepted in other journals |
| *Summer* | Guaranteed annual summer support in U.S. dollars |
| *Summer_yrs* | Number of years for which summer research support will last |
| *Budget* | Annual discretionary research budget |
| *Move Supp* | Moving support |
| *TenureReq_A* | Top-tier publication tenure requirements |
| *TenureReq_tot* | Total publication requirement for tenure at the offering school |

# Appendix B

## Time Complexity Analysis

**Table B1. Average Computation Time for the Simulated Dataset (in seconds)**

| mask | model | $n$ | $p$ | Benchmark | Random selection | Risk only | Weighted cost |
|---|---|---|---|---|---|---|---|
| BCADP | LM | 10,000 | 15 | 0.42 | 0.59 | 71.56 | 70.07 |
| | | | 20 | 0.50 | 0.81 | 118.50 | 116.24 |
| | | | 25 | 0.65 | 1.10 | 190.97 | 188.23 |
| | | 20,000 | 15 | 0.69 | 1.02 | 121.00 | 118.32 |
| | | | 20 | 0.93 | 1.20 | 222.27 | 217.05 |
| | | | 25 | 1.12 | 1.89 | 328.44 | 320.86 |
| | | 30,000 | 15 | 0.98 | 1.55 | 177.12 | 274.92 |
| | | | 20 | 1.32 | 1.69 | 306.43 | 295.94 |
| | | | 25 | 1.71 | 1.98 | 483.5 | 470.27 |
| | CART | 10,000 | 15 | 3.54 | 3.77 | 1172.85 | 1175.87 |
| | | | 20 | 4.39 | 4.75 | 1956.29 | 1960.87 |
| | | | 25 | 5.72 | 6.21 | 3146.46 | 3090.99 |
| | | 20,000 | 15 | 6.89 | 7.51 | 2231.46 | 2217.89 |
| | | | 20 | 10.32 | 11.00 | 4178.68 | 4344.34 |
| | | | 25 | 12.41 | 12.50 | 6097.34 | 5934.95 |
| | | 30,000 | 15 | 11.65 | 11.92 | 3429.32 | 3382.50 |
| | | | 20 | 15.91 | 15.85 | 6027.40 | 5826.10 |
| | | | 25 | 20.93 | 21.31 | 10040.04 | 9906.71 |
| MNS | LM | 10,000 | 15 | 0.46 | 0.61 | 84.71 | 81.45 |
| | | | 20 | 0.53 | 0.78 | 138.51 | 135.89 |
| | | | 25 | 0.70 | 0.94 | 225.68 | 220.84 |
| | | 20,000 | 15 | 0.78 | 1.01 | 142.55 | 190.62 |
| | | | 20 | 1.05 | 1.44 | 261.10 | 253.82 |
| | | | 25 | 1.28 | 1.61 | 387.46 | 377.60 |
| | | 30,000 | 15 | 1.15 | 1.45 | 208.78 | 201.91 |
| | | | 20 | 1.52 | 1.22 | 359.35 | 346.83 |
| | | | 25 | 2.27 | 1.48 | 569.98 | 554.85 |
| | CART | 10,000 | 15 | 3.29 | 3.38 | 1178.48 | 1182.15 |
| | | | 20 | 4.07 | 4.24 | 1942.91 | 1953.6 |
| | | | 25 | 5.41 | 5.63 | 3167.40 | 3117.33 |
| | | 20,000 | 15 | 6.26 | 6.69 | 2207.67 | 2193.92 |
| | | | 20 | 9.25 | 9.45 | 4136.88 | 4306.99 |
| | | | 25 | 11.29 | 10.9 | 6067.65 | 5934.00 |
| | | 30,000 | 15 | 10.48 | 9.97 | 3377.53 | 3337.82 |
| | | | 20 | 14.00 | 13.82 | 5946.59 | 5745.36 |
| | | | 25 | 17.76 | 18.02 | 9682.21 | 9626.88 |

**Note:** $n$ is the number of observations in the data set and $p$ is the number of features.

| Table B2. Average Computation Time for Three Real-World Datasets (in seconds) | | | | | | | | | | | | | |
|---|---|---|---|---|---|---|---|---|---|---|---|---|---|
| **Model** | **Metric** | **Credit** | | | | **Salary** | | | | **Compustat** | | | |
| | | ***bm*** | ***rnd*** | ***risk*** | ***wc*** | ***bm*** | ***rnd*** | ***risk*** | ***wc*** | ***bm*** | ***rnd*** | ***risk*** | ***wc*** |
| ***Masking Method: BCADP*** | | | | | | | | | | | | | |
| LM | R2 | 0.14 | 0.2 | 6.43 | 6.43 | 0.15 | 0.31 | 140.27 | 140.18 | 0.3 | 0.78 | 234.92 | 234.66 |
| | AAD | 0.14 | 0.2 | 6.52 | 6.54 | 0.15 | 0.31 | 141.78 | 141.84 | 0.3 | 0.78 | 238.23 | 238.09 |
| | RASD | 0.14 | 0.2 | 6.53 | 6.49 | 0.15 | 0.31 | 141.9 | 141.9 | 0.3 | 0.78 | 237.4 | 237.24 |
| CART | R2 | 0.44 | 0.51 | 17.18 | 17.09 | 0.27 | 0.44 | 343.66 | 343.71 | 4.03 | 5.42 | 4563.72 | 4555.76 |
| | AAD | 0.44 | 0.51 | 17.33 | 17.24 | 0.28 | 0.44 | 345.54 | 345.61 | 4.04 | 5.47 | 4628.34 | 4588.09 |
| | RASD | 0.44 | 0.51 | 17.27 | 17.21 | 0.27 | 0.44 | 345.52 | 345.61 | 4.02 | 5.42 | 4579.71 | 4561.62 |
| ***Masking Method: Laplace*** | | | | | | | | | | | | | |
| LM | R2 | 0.14 | 0.2 | 6.53 | 6.54 | 0.15 | 0.32 | 130.26 | 130.31 | 0.3 | 0.81 | 234.16 | 233.97 |
| | AAD | 0.14 | 0.2 | 6.5 | 6.51 | 0.15 | 0.32 | 131.73 | 131.74 | 0.3 | 0.82 | 237.36 | 237.26 |
| | RASD | 0.14 | 0.2 | 6.53 | 6.52 | 0.15 | 0.32 | 131.89 | 131.93 | 0.3 | 0.82 | 236.68 | 236.49 |
| CART | R2 | 0.44 | 0.51 | 17.17 | 17.14 | 0.27 | 0.45 | 333.34 | 333.55 | 4.05 | 5.43 | 4687.64 | 4630.56 |
| | AAD | 0.45 | 0.52 | 17.63 | 17.57 | 0.28 | 0.45 | 335.28 | 335.34 | 4.04 | 5.44 | 4715.5 | 4647.37 |
| | RASD | 0.44 | 0.51 | 17.35 | 17.32 | 0.28 | 0.45 | 335.39 | 335.44 | 4.05 | 5.44 | 4694.99 | 4629.02 |
| ***Masking Method: MNS2*** | | | | | | | | | | | | | |
| LM | R2 | 0.13 | 0.14 | 7.27 | 7.28 | 0.14 | 0.17 | 165.35 | 165.34 | 0.33 | 0.47 | 277.43 | 277.23 |
| | AAD | 0.13 | 0.14 | 7.32 | 7.33 | 0.14 | 0.17 | 166.88 | 167.01 | 0.34 | 0.47 | 280.68 | 280.49 |
| | RASD | 0.13 | 0.14 | 7.28 | 7.29 | 0.14 | 0.17 | 167.17 | 167.11 | 0.34 | 0.47 | 280.31 | 280.15 |
| CART | R2 | 0.47 | 0.49 | 19.74 | 19.73 | 0.29 | 0.32 | 402.89 | 402.92 | 4.55 | 4.76 | 5206.48 | 5207.77 |
| | AAD | 0.48 | 0.49 | 19.87 | 19.84 | 0.29 | 0.32 | 404.74 | 404.74 | 4.57 | 4.77 | 5223.29 | 5230.29 |
| | RASD | 0.47 | 0.49 | 19.79 | 19.79 | 0.29 | 0.32 | 405.3 | 405.04 | 4.52 | 4.73 | 5198.14 | 5199.38 |
| ***Masking Method: TwinUnif*** | | | | | | | | | | | | | |
| LM | R2 | 0.13 | 0.14 | 7.32 | 7.33 | 0.14 | 0.16 | 165.23 | 165.2 | 0.33 | 0.43 | 275.77 | 275.67 |
| | AAD | 0.13 | 0.14 | 7.31 | 7.32 | 0.14 | 0.16 | 166.95 | 166.97 | 0.33 | 0.43 | 279.41 | 279.31 |
| | RASD | 0.13 | 0.14 | 7.31 | 7.3 | 0.14 | 0.16 | 167.12 | 167.09 | 0.33 | 0.43 | 278.73 | 278.58 |
| CART | R2 | 0.45 | 0.45 | 19.06 | 19.08 | 0.28 | 0.31 | 403.62 | 403.54 | 4.65 | 4.93 | 5331.48 | 5320.43 |
| | AAD | 0.46 | 0.46 | 19.26 | 19.26 | 0.29 | 0.31 | 405.47 | 405.39 | 4.61 | 4.89 | 5322.14 | 5321.13 |
| | RASD | 0.45 | 0.46 | 19.29 | 19.29 | 0.29 | 0.31 | 405.56 | 405.54 | 4.6 | 4.89 | 5308.02 | 5300.69 |

**Note:** *bm* is the benchmark method selection method, where only the confidential feature is masked; *rnd* is the selection method, where the confidential feature and a randomly selected half of the nonconfidential features are selected for masking; *risk* is the risk-only feature selection method; and *wc* is the weighted-cost feature selection method.